\documentclass{article}

\usepackage[final, nonatbib]{neurips_2020}
\usepackage[utf8]{inputenc} 
\usepackage[T1]{fontenc}    
\usepackage{hyperref}       
\usepackage{url}            
\usepackage{booktabs}       
\usepackage{amsfonts}       
\usepackage{nicefrac}       
\usepackage{microtype}      
\usepackage{multirow}
\usepackage{multicol}
\usepackage[dvipsnames]{xcolor}
\usepackage{amsmath}
\usepackage{amsthm}
\usepackage{graphicx}
\usepackage{color}
\usepackage[utf8]{inputenc}
\usepackage{appendix}
\usepackage{thmtools,thm-restate}
\usepackage{caption}
\usepackage{subcaption}
\usepackage{wrapfig}
\usepackage{lipsum}
\usepackage{enumitem}
\usepackage{xcolor}
\usepackage[sorting=none, style=numeric-comp]{biblatex}
\usepackage[ruled,linesnumbered]{algorithm2e}
\SetArgSty{textup}
\SetKw{Continue}{continue}

\newtheorem{prop}{Proposition}

\newtheorem{cor}{Corollary}

\newtheorem{lm}{Lemma}

\newtheorem{thm}{Theorem}

\newcommand{\be}{\begin{eqnarray}}
\newcommand{\ee}{\end{eqnarray}}
\newcommand{\benn}{\begin{eqnarray*}}
\newcommand{\eenn}{\end{eqnarray*}}
\def\IR{\rm I \kern-0.20em R}

\newcommand{\bthm}{\begin{thm}}
\newcommand{\ethm}{\end{thm}}
\newcommand{\bcor}{\begin{cor}}
\newcommand{\ecor}{\end{cor}}
\newcommand{\bprop}{\begin{prop}}
\newcommand{\eprop}{\end{prop}}
\newcommand{\blm}{\begin{lm}}
\newcommand{\elm}{\end{lm}}
\newcommand{\beq}{\begin{equation}}
\newcommand{\eeq}{\end{equation}}
\newcommand{\ber}{\begin{eqnarray}}
\newcommand{\eer}{\end{eqnarray}}
\newcommand{\bproof}{\begin{proof}}
\newcommand{\eproof}{\end{proof}}

\newcommand{\bit}{\begin{itemize}}
\newcommand{\eit}{\end{itemize}}
\newcommand{\ben}{\begin{enumerate}}
\newcommand{\een}{\end{enumerate}}
\newcommand{\bdesc}{\begin{description}}
\newcommand{\edesc}{\end{description}}
\newcommand{\beqarrn}{\begin{eqnarray*}}
\newcommand{\eeqarrn}{\end{eqnarray*}}
\newcommand{\bproofof}{\begin{proofof}}
\newcommand{\eproofof}{\end{proofof}}
\newenvironment{rem}{\begin{trivlist}\item[]{\bf
Remark:}\hspace{4mm}}{\end{trivlist}}
\newcommand{\brem}{\begin{rem}}
\newcommand{\erem}{\end{rem}}
\newenvironment{rems}{\begin{trivlist}\item[]{\bf
Remarks}\begin{itemize}}{\end{itemize}\end{trivlist}}
\newcommand{\brems}{\begin{rems}}
\newcommand{\erems}{\end{rems}}
\newtheorem{fact}{Fact}
\newcommand{\bfact}{\begin{fact}}
\newcommand{\efact}{\end{fact}}
\newtheorem{examp}{Example}
\newcommand{\bexamp}{\begin{examp}\rm}
\newcommand{\eexamp}{\end{examp}}
\newtheorem{defn}{Definition}
\newcommand{\bdefn}{\begin{defn}\rm}
\newcommand{\edefn}{\end{defn}}

\newtheorem{alg}{Algorithm}
\newcommand{\balg}{\begin{alg}}
\newcommand{\ealg}{\end{alg}}
\newtheorem{prob}{Problem}
\newcommand{\bprob}{\begin{prob}}
\newcommand{\eprob}{\end{prob}}

\newcommand{\bvtm}{\begin{verbatim}}
\newcommand{\bfig}{\begin{figure}}
\newcommand{\efig}{\end{figure}}
\newcommand{\bcen}{\begin{center}}
\newcommand{\ecen}{\end{center}}

\long\def\comment#1{}

\def \n2{{N_0 \over 2}}

\def \h5{\hspace{0.5in}}

\newcommand{\qedwhite}{\hfill \ensuremath{\Box}}

\title{Towards Interaction Detection Using Topological Analysis on Neural Networks}

\author{%
  Zirui Liu\\
  Dept. of Computer Science\\
  Texas A\&M University\\
  College Station, TX \\
  \texttt{tradigrada@tamu.edu} \\
   \And
  Qingquan Song\\
  Dept. of Computer Science\\
  Texas A\&M University\\
  College Station, TX \\
   \texttt{song\_3134@tamu.edu} \\
   \And
  Kaixiong Zhou\\
  Dept. of Computer Science\\
  Texas A\&M University\\
  College Station, TX \\
   \texttt{zkxiong@tamu.edu} \\
   \AND
  Ting-Hsiang Wang\\
  Dept. of Computer Science\\
  Texas A\&M University\\
  College Station, TX \\
   \texttt{thwang1231@tamu.edu} \\
   \And
  Ying Shan\\
  Tencent Company\\
  Beijing, China\\
   \texttt{yingsshan@tencent.com} \\
    \And
  Xia Hu\\
  Dept. of Computer Science\\
  Texas A\&M University\\
  College Station, TX \\
   \texttt{hu@cse.tamu.edu} \\
}

\newtheorem{corol}{Corollary}
\newtheorem{lemma}{Lemma}
\newtheorem{definition}{Definition}

\begin{document}

\maketitle
\begin{abstract}
Detecting statistical interactions between input features is a crucial and challenging task. Recent advances demonstrate that it is possible to extract learned interactions from trained neural networks. It has also been observed that, in neural networks, any interacting features must follow a strongly weighted connection to common hidden units. Motivated by the observation, in this paper, we propose to investigate the interaction detection problem from a novel topological perspective by analyzing the connectivity in neural networks. Specially, we propose a new measure for quantifying interaction strength, based upon the well-received theory of persistent homology. Based on this measure, a \textbf{P}ersistence \textbf{I}nteraction \textbf{D}etection~(PID) algorithm is developed to efficiently detect interactions. Our proposed algorithm is evaluated across a number of interaction detection tasks on several synthetic and real world datasets with different hyperparameters. Experimental results validate that the PID algorithm outperforms the state-of-the-art baselines.
\end{abstract}

\section{Introduction}


Statistical interaction describes a subset of input features that interact with each other to have an effect on outcomes.
For example, using Phenelzine together with Fluoxetine may lead to serotonin syndrome \cite{sun2008drug}.
Interaction detection problem is to quantify the influence of any subset of input features that may potentially be an interaction. The quantified influence in the problem is called interaction strength. With detected interactions, we may formulate hypotheses that could lead to new data collection and experiments. 
Traditional methods often need to conduct individual tests for all interaction candidates~\cite{fisher1992statistical, sorokina2008detecting} or pre-specify all functional forms of interests~\cite{bien2013lasso,tibshirani1996regression}. Recent efforts have been dedicated to extracting learned interactions in neural networks by designing measures for quantifying interaction strength based on predefined conditions in a heuristic way~\cite{NID}.
It has been shown to be an effective way to detect interactions and avoid the drawbacks of traditional methods.

One key observation in the state-of-the-art methods is that any interacting features must follow strongly weighted connections to a common hidden unit before reaching the final output layer~\cite{NID,NIT}. Based on this, the strength of interactions can be modeled by the connectivity between these interacting features and output units of a trained neural network. This motivates us to solve the problem from a 
novel topological perspective. Specifically, our framework builds upon computational techniques from algebraic topology, specially the persistent homology, which has been shown beneficial for several deep learning models~\cite{hofer2017deep,khrulkov2018geometry,hofer2019learning}. The main advantages of utilizing persistent homology are twofold. First, it provides us a rigorous mathematical framework for analyzing the connectivity in a trained neural network. Second, persistent homology can be used to quantify the importance of each connected component in the neural network, and the connectivity between interacting features and output units are characterized by these connected components.

However, persistent homology cannot be directly applied to quantify the interaction strength from the importance of connected components that link interacting features to  units in the final output layers. Also, an interaction is a subset of input features, which is not within the scope of 
persistent homolgy. The key challenge remains to define a measure for quantifying the 
interaction strength, which should provide meaningful insights while maintaining theoretical generality. 

In this paper, we show that the key concepts of \textit{persistence diagrams} in persistent homology theory can be extended to interactions for tackling the challenges. Specifically, we propose a new measure for quantifying interaction strength, which is computed to reflect the connectivity between interacting features and output units in a neural network. Based on the measure, we propose Persistence Interaction Detection~(PID), a framework that can efficiently extract interactions from neural networks. We also prove that our framework is locally stable, meaning that PID is not sensitive to the perturbation of weights in neural networks. Formally, our contributions are as follows:
\begin{itemize}[leftmargin=*]
\setlength\itemsep{5pt}
\item We formulate the interaction detection problem as a topology problem. Based on the persistent homology theory, we propose a new measure for quantifying interaction strength by analyzing the topology of neural networks. We then provide analysis for the measure from different perspectives.
\item We derive an efficient algorithm to calculate the proposed interaction strength measure. Also, we theoretically analyze the local stability of our proposed framework.
\item The proposed PID framework demonstrates strong performance across different tasks, network architectures, hyperparameter settings, and datasets.
\end{itemize}

\section{Preliminaries}
\label{sec: prelim}
We first introduce the notations and give the formal definition of feature interactions. Based on the notations, we introduce concepts of the filtration and persistence diagrams. We then show how to build filtration for neural networks in~\ref{sec:background}, serving as the preliminary of our proposed method.

\subsection{Problem Formulation and Notations}
\label{subsec: notations}
We denote vectors with boldface lowercase letters~(e.g., \textbf{x}, \textbf{w}), matrices with boldface capital letters~(e.g. \textbf{W}), and scalars with lowercase letters~(e.g., $a$). We use $x_{i}$ to represent the $i$-th entry of vector \textbf{x}, and $W_{ij}$ to denote the entry in the $i^{\mathrm{th}}$ row and $j^{\mathrm{th}}$ column of $\mathbf{W}$. The transpose of a matrix or a vector is denoted as $\mathbf{W}^{\top}$ or $\mathbf{x}^{\top}$. For a set $\mathcal{S}$, its cardinality is denoted by $|\mathcal{S}|$. We use $\mathcal{S}\backslash i$ to denote the set $\{j|j\in\mathcal{S} ~\text{and}~ j \neq i\}$. Let $\mathbf{x}\in\mathbb{R}^{d}$ be the feature vector. An interaction $\mathcal{I}$ is a set of interacting features, where $|\mathcal{I}| \geq 2$. 
A $K$-order interaction $\mathcal{I}$ satisfies $|\mathcal{I}|=K$. A high-order interaction is an interaction whose order $\geq 3$. We will write $\textbf{x}^{\mathcal{I}}\in\mathbb{R}^{|\mathcal{I}|}$ as the feature vector selected by $\mathcal{I}$.

Consider a feed-forward neural network~(FNN) with $L$ hidden layers~(e.g., an MLP). Let $p_{l}$ be the number of hidden units at the $l^{\mathrm{th}}$ layer. The input features are treated as the $0^{\mathrm{th}}$ layer and $p_{0}=d$ is the number of input features. The $l^{\mathrm{th}}$ layer weight matrix is denoted by $\textbf{W}^{(l)}\in \mathbb{R}^{p_{l-1}\times p_{l}}$. Given a FNN with weights $\{\textbf{W}^{(i)}\}_{i=1}^{L}$, 
its equivalent weighted directed acyclic graph $\mathcal{G}(V,E)$ can be constructed as follows: 
We create a vertex for each hidden unit in the neural network and consequently the set of all vertices: $V=\{v_{l,i}|\forall~l,i\}$, 
where $v_{l,i}$ represents the $i^{\mathrm{th}}$ hidden unit at the $l^{\mathrm{th}}$ layers; and
we assign weight $W^{(l)}_{i,j}$ to each edge in $E=\{(v_{l-1,i}, v_{l,j})|\forall l, i,j\}$.

In this work, we focus on detecting non-additive interactions. The non-additive interaction is formally defined in Definition~\ref{def:FID}. We remark that detecting ``additive interactions'' is a trivial task because any ``additive interactions'' can be decomposed to the sum of two terms, and non-additive interactions are those which cannot be further decomposed.

\begin{definition}[Non-additive interactions~\cite{sorokina2008detecting,friedman2008predictive}]
\label{def:FID}
Let $\{0,...,d-1\}$ denotes the input feature set. Given a function $f$: $\mathbb{R}^{d}\to\mathbb{R}$ and a feature vector $\mathbf{x}=(x_{1},...,x_{d})$, 
$f$ shows no non-additive interaction of $\{x_{i}$,$x_{j}\}$ if $f$ can be expressed as the sum of two functions, $f_{\backslash i}$ and $f_{\backslash j}$, 
where $f_{\backslash i}$ is a function which does not depend on $x_{i}$ and $f_{\backslash j}$ is a function which not depend on $x_{j}$:
\be
f(\mathbf{x})=f_{\backslash i}(\mathbf{x}^{\{0,...,d-1\}\backslash i}) + 
f_{\backslash j}(\mathbf{x}^{\{0,...,d-1\}\backslash j}). \nonumber
\ee
\end{definition}

For example, in the function of $\pi^{x_{0}x_{1}}+\log(x_{1}+x_{2}+x_{4})$, there is a pairwise interaction $\{0, 1\}$ and a 3-order interaction $\{x_1,x_2,x_4\}$. In contrast, $\{x_0,x_1,x_2,x_4\}$ is a spurious interaction. The goal of interaction detection algorithms is to map models into a set of their learned interaction candidates associated with interaction strength. Ideally, a larger value of interaction strength should  indicate the true interaction instead of a spurious interaction.

\subsection{Persistent Homology on Neural Networks}
\label{sec:background}

Persistent homology is an algebraic method for identifying the most prominent connectivity characterizing a geometric object, which is widely used in medical imaging and geometric modeling \cite{edelsbrunner2005surface,roerdink2000watershed}. In this paper, the object we studied is a weighted directed graph $\mathcal{G}(V, E)$ corresponding to a trained feed-forward neural network. In topology, connected components represent the connectivity of the graph. We can apply persistent homology theory to quantify the importance of each connected component in $\mathcal{G}$. To be specific, $(\mathcal{G}, \phi)$ is called a \textit{size pair}~\cite{d2010natural}, where $\phi$ is a \textit{measureing function}~\cite{frosini1999size}. The role of $\phi$ is to take into account the connective properties of $\mathcal{G}$. The $\lambda$-threshold set of $(\mathcal{G}, \phi)$ is defined as follows:
\be
L^{\lambda}=\{x|x \in E, \phi(x)\geq\lambda\}, \nonumber
\ee
where $x$ is the edge of $\mathcal{G}$. The measuring function $\phi:E\rightarrow\mathbb{R}$ maps a specific edge to a real number.

\begin{definition}[Filtration~\cite{edelsbrunner2010computational}]
\label{def:filtration}
Without loss of generality, suppose $\lambda_1 >\lambda_2$, if the corresponding threshold sets satisfy $L^{\lambda_1} \subseteq L^{\lambda_2}$, then $\phi$ is non-decreasing over $\mathcal{G}$.
Given $(\mathcal{G}, \phi)$, where $\phi$ is non-decreasing over $\mathcal{G}$, and a set of thresholds follow $\lambda_{0}\geq \lambda_{1}\geq...\geq\lambda_{n}$, the collections of threshold sets $L^{\lambda_{0}}\subseteq L^{\lambda_{1}}\subseteq...\subseteq L^{\lambda_{n}}$ is called a \textbf{filtration} of $(\mathcal{G},\phi)$.
\end{definition}

We propose to build the filtration for FNNs and define the measuring function as follows. Let $\mathcal{W}$ be the set of weights. 
Given $\mathcal{W}$ of a trained feed-forward neural network such that $w_{max}:=\textrm{max}_{w\in\mathcal{W}}|w|$ 
and $\mathcal{W'}:=\{|w|/w_{max}|w\in\mathcal{W}\}$, 
where $\mathcal{W'}$ is indexed in non-ascending order, namely $1=w'_{0}\geq w'_{1}\geq...\geq w'_{n}\geq0$. The weights associated with edges reflect the connectivity between vertices in the networks. Similar to \cite{rieck2018neural}, the measuring function $\phi$ for $\mathcal{G}$ is defined as 
$\phi((v_{l-1,i},v_{l,j}))=|W^{(l)}_{i,j}|/w_{max}, \forall(v_{l-1,i},v_{l,j})\in E$, which represents the edge strength. The sorted weights are used as $\lambda$ in Definition~\ref{def:filtration}. Consequently, the filtration can be constructed as $\mathcal{G}^{w'_{0}} \subseteq \mathcal{G}^{w'_{1}} \subseteq...$, 
where $\mathcal{G}^{w'_{i}}=(V, \{(u,v)|(u,v)\in E \wedge \phi((u,v))\geq w'_{i} \})$. We remark that $\mathcal{G}^\lambda$ is both a subgraph of $\mathcal{G}$ and a $\lambda$-threshold set of size pair $(\mathcal{G},\phi)$. $\mathcal{G}^{w'_0}$ is the sub-graph with exact one edge which has greatest weight. As shown in Figure~\ref{fig:illu of filtrations}, when the thresholds are decreased, edges are added into the sub-graph and vertices will be connected. It is summarized in Figure~\ref{fig:illu of filtrations}.

The interpretation of $\mathcal{G}^\lambda$ is that, $\mathcal{G}^\lambda$ is the image of $\mathcal{G}$ at different spatial resolution. Edges with larger weights, which indicates stronger connectivity, will appear over a wide range of spatial scales. As the threshold of filters decreases, edges with smaller weights, which indicates weaker connectivity, will start to pass the filter and provide detailed information of $\mathcal{G}$. In the filtration process, these gradually added edges will form different connected components. From persistent homology theory, persistent connected components, which are detected over a wide range of spatial scales, are more representative for the connectivity 
pattern of $\mathcal{G}$~\cite{TopologyAndData}. Based on this, persistence diagram is a computational tool for quantifying the importance of these emerged connected components.

Given the size pair $(\mathcal{G}, \phi)$, when we decrease $\lambda$, connected components can be \textit{created} (new edges are added, forming new components) or \textit{destroyed} (two connected components joining together). 
For each connected component $i$, the threshold causes the birth of $i$ is called the \textit{birth} time $b_{i}$ 
and the threshold causes the death of $i$ is called the \textit{death} time $d_{i}$. 
The persistence diagram tracks these changes and represents creation and destruction of $i$ as a tuple ($b_{i}, d_{i}$). It quantifies the importance of each connected component by its lifetime~(persistence).

\textbf{Persistence Diagrams}\quad Given the filtration of a size pair $(\mathcal{G},\phi)$, with the \textit{birth} time $b_{i}$ and the \textit{death} time $d_{i}$ of each connected component $i$ appearing in the filtration, the collection of the \textit{birth} time and the \textit{death} time tuple $\mathcal{D}=\{(b_{i}, d_{i})|\forall i$ appears in the filtration$\}$ is called the \textbf{persistence diagrams} of $(\mathcal{G},\phi)$. The \textbf{persistence} of $i$ is $\mathrm{per}(i)=|b_{i}-d_{i}|$.

\begin{figure}[hbt!]
  \centering
  \includegraphics[scale=0.30]{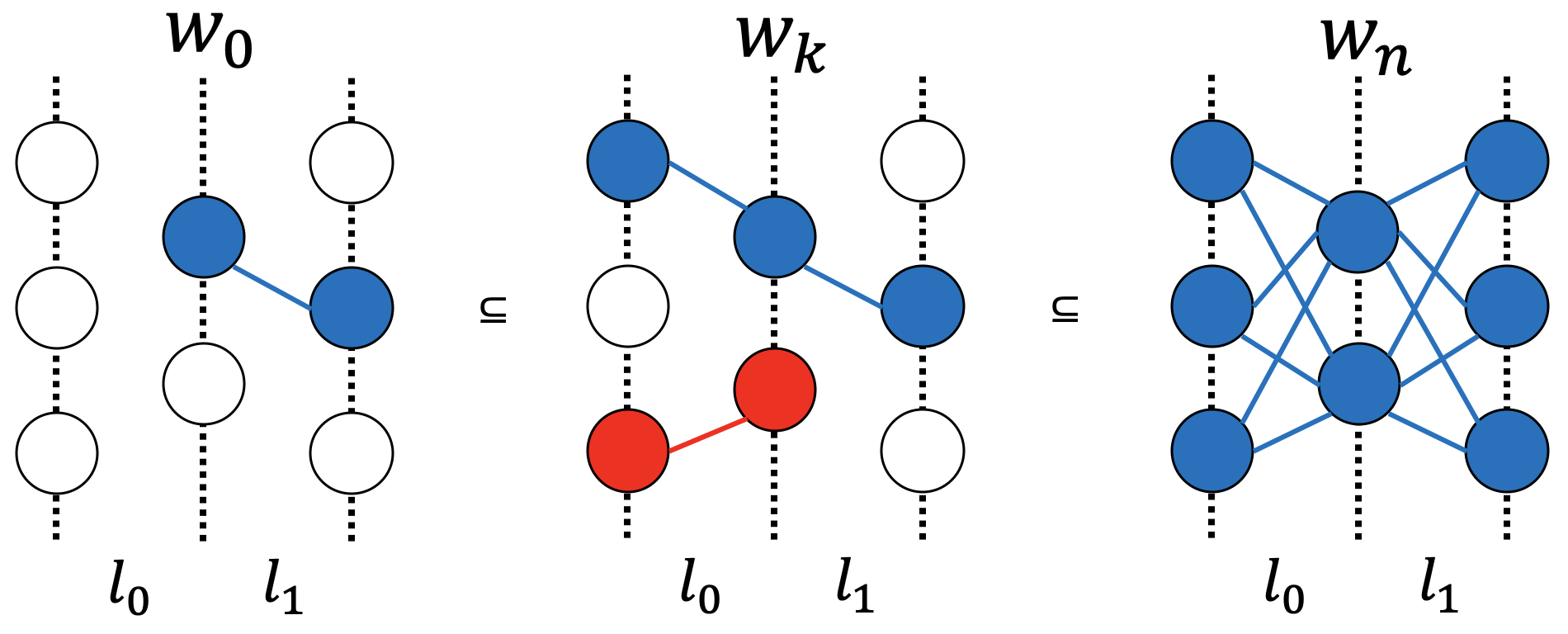}
  \caption{The filtration of a network with two layers. The color scheme illustrates the connected components. The filtration process is represented by colouring
connected components that are created or merged when the respective weights are greater than or equal to the threshold $w_i$.}
  \label{fig:illu of filtrations}
\end{figure}

\section{Persistence Interaction Detection}
\label{sec:pd}
In this section, we present the proposed PID framework for detecting feature interactions in neural networks. The key intuition of our proposed method is to formulate the interaction detection as a topology problem. That is to find the long-lived connectivity between interactions and outputs in the neural network over a wide range of spatial scales. Based on this, we propose a new measure for quantifying interaction strength~(section~\ref{subsec:pd of feature subset}). Subsequently, we derive our proposed Persistence Interaction Detection algorithm for calculating the measure efficiently~(section \ref{subsec: pid}). We then give the stability analysis for our proposed algorithm~(section \ref{subsec: stab of pid}). To avoid creating confusion in terminology, by ``interaction'', we mean a subset of input features that satisfies Definition~\ref{def:FID} in the rest of the paper.



\subsection{Persistence As a Measure For Interaction Strength}  
\label{subsec:pd of feature subset}
The concepts of the \textit{birth} time and the \textit{death} time are originally defined for connected components in persistent homology theory. However, the persistence of connected components only implies the importance of themselves. For a particular interaction $\mathcal{I}$, we cannot directly obtain the importance of $\mathcal{I}$ from the persistence of connected components that contain $\mathcal{I}$. Also, an interaction $\mathcal{I}$ is a subset of input features that is not within the scope of 
persistent homolgy. In this subsection, we will extend these concepts to interactions for deriving a new measure for quantifying interaction strength.

  From Definition \ref{def:FID}, an interaction is a set of associate features that have an effect on the output. Inspired by persistent homology, we can model this effect by the connectivity between interactions and output units in neural networks. Informally, the \textit{birth} time of an interaction is when there exists a path connecting it to the final output layer, and the \textit{death} time is when the path is also connected to any additional input feature in the filtration. After extending concepts of the $birth$ time and the $death$ time to interactions,  we can obtain persistence diagrams of interactions and the interaction strength can be quantified from the lifetime of the connectivity. We first give the definition for the connectivity strength between the interactions and the units in the final output layer in Definition~\ref{def:connected_path}. Based on the quantified connectivity, we formally define the \textit{birth} time and the \textit{death} time of an interaction.

\begin{definition}[$\langle\phi=\lambda\rangle$-connected]
\label{def:connected_path}
 Let $(\mathcal{G}, \phi)$ be the corresponding size pair and $\mathcal{G}^{w'_{0}} \subseteq \mathcal{G}^{w'_{1}} \subseteq...$ be the filtration of a neural network, respectively;
 and $\{0,...,d-1\}$ denotes the set of input features. 
 For a feature subset $\mathcal{I}$ and a real-number threshold $\lambda$, we call $\mathcal{I}$ and the final output units are $\langle\phi=\lambda\rangle$-connected if: first, there exists a connected component $A \subseteq \mathcal{G}^{\lambda}$ containing $\mathcal{I}$ and the final output units;
 second, for any such connected component $A$, $\forall i\in\{0,...,d-1\}\backslash \mathcal{I}\}$, it satisfies $i \notin A$.
\end{definition}


\textbf{Persistence diagrams of interactions} \quad Given the threshold $\lambda_{b}$, suppose: The feature subset $\mathcal{I}$ and the final output unit are $\langle\phi=\lambda_{b}\rangle$-connected and, $\forall\lambda_{i}\geq\lambda_{b}$, 
$\mathcal{I}$ and the final output unit are not $\langle\phi=\lambda_{i}\rangle$-connected, then we call $\lambda_{b}$ the \textit{birth} time of $\mathcal{I}$. Correspondingly, 
the \textit{death} time $\lambda_{d}$ of $\mathcal{I}$ is that $\forall\lambda_{i}\leq\lambda_{d}$, $\mathcal{I}$ and the outputs become not $\langle\phi=\lambda_{i}\rangle$-connected, i.e., interaction $\mathcal{I}$ no longer exists due to the addition of other input features. The collection of the \textit{birth} time and the \textit{death} time tuple $\mathcal{D}=\{(b_{\mathcal{I}}, d_{\mathcal{I}})|\forall \mathcal{I} \subseteq \{0,...,d-1\}$\} is called the \textbf{persistence diagrams} of interactions.

After defining the \textit{birth} time and the \textit{death} time of an interaction $\mathcal{I}$, we can quantify its interaction strength by its persistence. We remark that the aforementioned process creates new interaction candidates by associating new features with existing interaction candidates. Some interaction candidates might never born. An example of the persistence of interactions is illustrated in Figure~\ref{fig:illu_inter_pd}. 
\begin{wrapfigure}{r}{0.48\textwidth}
    \centering
    \includegraphics[width=0.46\textwidth]{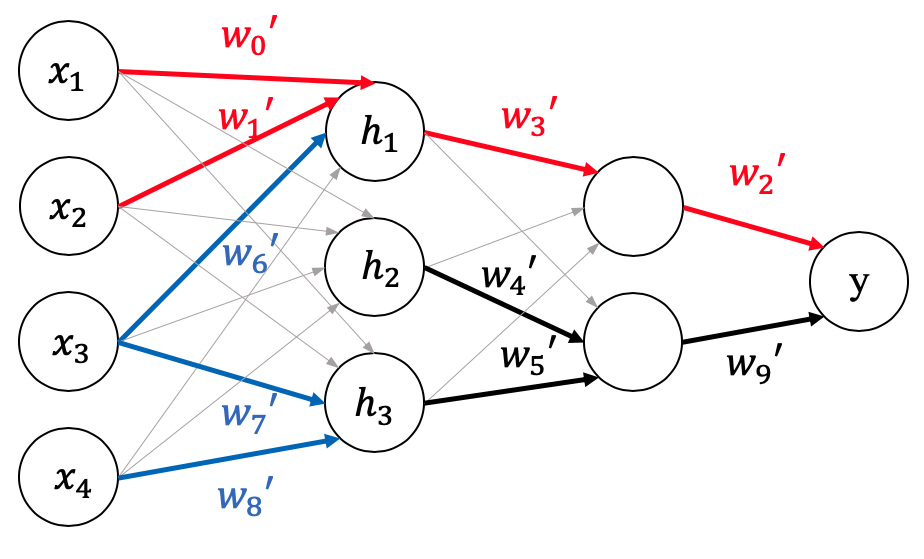}
    \caption{An example for illustrating persistence of interactions.}
    \label{fig:illu_inter_pd}
\end{wrapfigure}
Let $\mathbf{x}=[x_1,x_2,x_3,x_4]$. $y=x_1^{x_2}+\frac{x_3 x_4}{1000}$. We train a neural network $f$ to minimize the loss $\mathcal{L}(f(x),y)+\mathcal{R}(f)$, where $\mathcal{L}$ is the mean square error and $\mathcal{R}$ is the regularization term. Suppose $w'_0\geq...\geq w'_9$ are the top ten largest weights in $\mathcal{W'}$. Then the interaction $\{x_1, x_2\}$ and $y$ are $\langle\phi=w'_3\rangle$-connected because of the connected component marked in red. The birth time and the death time of $\{x_1, x_2\}$ are $w'_3$ and $w'_6$, respectively. And the death time of  $\{x_1, x_2\}$ marks the birth time of $\{x_1,x_2,x_3\}$.



Intuitively, strength can be quantified in terms of the minimal ``amount of change'' necessary to eliminate a learned interaction in NNs. The ``amount of change'' is referred to the distance between changed weights and original weights. From this perspective, the proposed measure is the minimal ``amount of change'' to eliminate the $\langle\phi=\lambda\rangle$-connectivity between interactions and outputs. For example, if we change $w'_6$ to be as large as $w'_1$, then the input feature $x_2$ and $x_3$ will be simultaneously added to $\{x_1\}$ to form $\{x_1,x_2,x_3\}$, thus $\{x_1, x_2\}$ will never born. The persistence of $\{x_1,x_2\}$ is the gap between the red connected component's smallest weight and $w'_6$~(i.e., $|w'_3-w'_6|$). 
This gap is the minimal amount of change to eliminate the $\langle\phi= w'_3\rangle$-connectivity between $\{x_1,x_2\}$ and $y$.

\subsection{Ranking Interactions Using PID}
\label{subsec: pid}
In this subsection we present our PID framework that calculates the proposed measure efficiently. To detect these $\langle\phi=\lambda\rangle$-connectivity between interactions and outputs in Definition~\ref{def:connected_path}. We define the mask matrix $\mathbf{M}^{\lambda}_{(l)}\in\mathbb{R}^{p_{l-1}\times p_l}$ for the $l^{\mathrm{th}}$ layer as
 \be
 [\mathbf{M}^{\lambda}_{(l)}]_{i,j}=\left\{\begin{array}{ccl}
      & 1, & \quad\mathrm{if}\quad\phi((v_{l-1,i},v_{l,j}))\geq\lambda.\\
      & 0, & \quad \mathrm{otherwise}.
 \end{array}\right.
 \ee
The aggregated mask matrix $\mathbf{M}^{\lambda}\in\mathbb{R}^{p_L \times d}$ are defined as:
 \be
 \mathbf{M}^{\lambda}=(\mathbf{M}^{\lambda}_{(L)})^{\top}\cdot(\mathbf{M}^{\lambda}_{(L-1)})^{\top}\cdot\cdot\cdot(\mathbf{M}^{\lambda}_{(1)})^{\top}.
 \ee

\begin{restatable}[Proof in Appendix~\ref{appendix:proof for lem 1}]{lemma}{lemmamask}
\label{lemma:mask}
Let $\{0,...,d-1\}$ denotes the input feature set, and $\mathbf{M}^{\lambda}$ denotes the aggregated mask matrix corresponding to threshold $\lambda$, where the $r^{\mathrm{th}}$ row of $\mathbf{M}^{\lambda}$ is denoted as $\mathbf{m}^{\lambda}_{r}\in\mathbb{R}^{d}$. The feature subset $\mathcal{I}$ and the corresponding $r^{th}$ unit at the final output layer are $\langle\phi=\lambda\rangle$-connected if all elements in $[\mathbf{m}^{\lambda}_{r}]^{\mathcal{I}}\in \mathbb{R}^{|\mathcal{I}|}$ are non-zero and all other elements in $[\mathbf{m}^{\lambda}_{r}]^{\{0,...,d-1\}\backslash \mathcal{I}}$ are zero, where $[\mathbf{m}^{\lambda}_{r}]^{\mathcal{I}}$ is the subvector of $\mathbf{m^\lambda_r}$ selected by $\mathcal{I}$.
\end{restatable}

As pointed out in~\cite{lecun2015deep}, different neurons are activated by different patterns~(patterns are exactly interactions of raw input features). This indicates that we should generate interaction candidates for each neuron separately. With Lemma~\ref{lemma:mask}, we can detect the $\langle\phi=\lambda\rangle$-connectivity between interactions and units in the output layer. 
 However, only care the $\langle\phi=\lambda\rangle$-connectivity between them 
will ignore the difference between neurons. For example, in Figure~\ref{fig:illu_inter_pd}, all neurons share common interaction candidates in 
the aforementioned process. The edges gradually added by the filtration process sequentially create the interaction candidates $\{x_1,x_2\}$, $\{x_1,x_2,x_3\}$ and $\{x_1,x_2,x_3,x_4\}$. $\{x_3,x_4\}$ will not be considered because $x_3$ has been merged with $\{x_1,x_2\}$ when they meet at $h_1$. But clearly, $x_3$ and $x_4$ 
might be a potential interaction candidate because the activation pattern of $h_3$ is largely determined by $x_3$ and $x_4$. To generate interaction candidates for each neuron at a particular layer $l$, we decompose $\mathbf{M}^{\lambda}$ into  
$\mathbf{M}^{\lambda_{up}}_{(l)}\in\mathbb{R}^{p_L\times p_{l}}$ and $\textbf{M}^{\lambda_{down}}_{(l)}\in\mathbb{R}^{p_{l}\times d}$, where
\be
\label{eq:up_down}
\left\{\begin{array}{ccl}
\mathbf{M}^{\lambda_{up}}_{(l)}=(\mathbf{M}^{\lambda}_{(L)})^{\top}\cdot\cdot\cdot(\mathbf{M}^{\lambda}_{(l)})^{\top}.\\
\mathbf{M}^{\lambda_{down}}_{(l)}=(\mathbf{M}^{\lambda}_{(l-1)})^{\top}\cdot\cdot\cdot(\textbf{M}^{\lambda}_{(1)})^{\top}.
\end{array}\right.
\ee

\begin{wrapfigure}{r}{0.35\textwidth}
    \centering
    \includegraphics[width=0.32\textwidth]{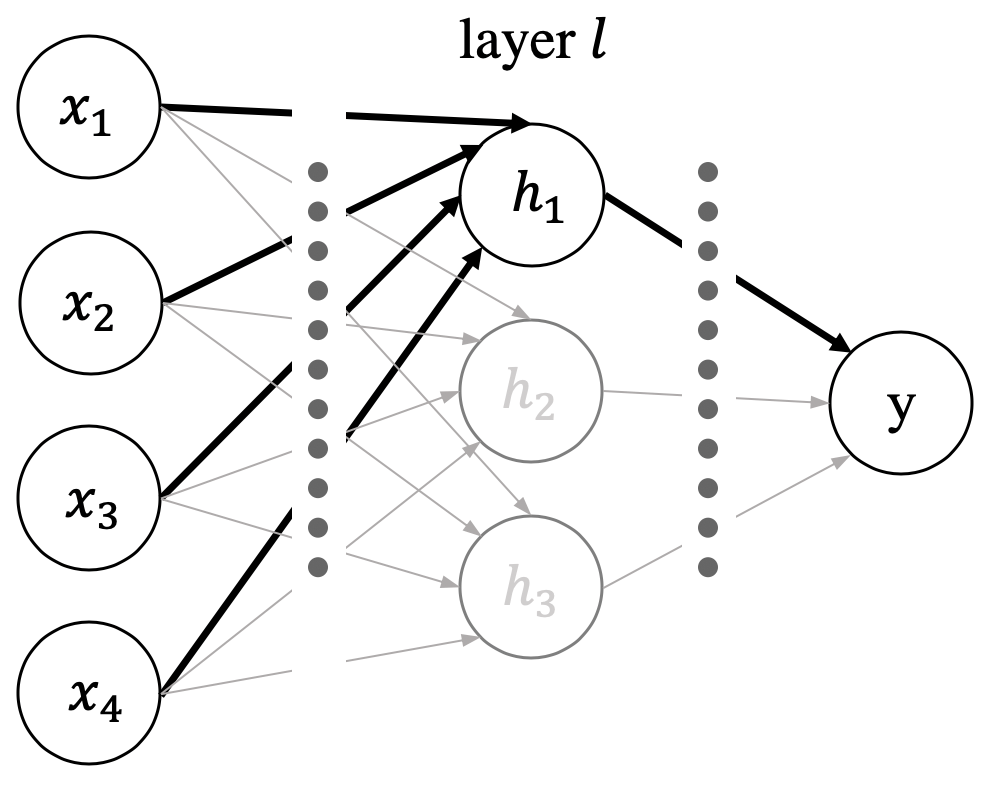}
    \caption{Illustration of PID.}
    \label{fig:illu of pid}
\end{wrapfigure}

We can obtain the connectivity between a particular neuron $r$ at layer $l$ and units in the final output layer from $\mathbf{M}^{\lambda_{up}}_{(l)}$~(by viewing the layer $l$ as the input layer in the Lemma~\ref{lemma:mask}). Similarly, the connectivity between the neuron $r$ and input features can be inferred from  $\mathbf{M}^{\lambda_{down}}_{(l)}$. For each neuron $r$ at layer $l$, we generate interaction candidate $\mathcal{I}$ for $r$ only if: first, $\mathcal{I}$ and $r$ are connected; second, $r$ and units in output layer are connected. It is summarized in Figure \ref{fig:illu of pid}. For example, in Figure \ref{fig:illu_inter_pd}, if the layer $l$ is set to the first layer, then PID will first generate $\{x_3,x_4\}$ for $h_3$ because $\{x_3,x_4\}-h_3-y$ are connected once the threshold achieves $w'_9$. For the same interaction candidate generated at different neurons, we aggregate their persistence. We list the full algorithm in Appendix~\ref{appendix: full PID algo}. From the topological perspective, we model how interactions influence a particular neuron at layer $l$, as well as how this neuron influences units in the final output layer by the quantified connectivity between them. $p$ in algorithm~\ref{algo:id} is the norm of the persistence diagram. The $p$-norm is known to be a stable summary for persistence diagrams~\cite{cohen2010lipschitz}.

\subsection{Stability of Persistence Interaction Detection}
\label{subsec: stab of pid}
An interaction detection algorithm should not be sensitive to the perturbation of weights, e.g., training neural networks with one or two extra epochs should only change the proposed interaction strength a little. We call that these insensitive algorithms are locally stable. It should be noted that local stability is a necessary condition for fidelity. If the algorithm gives totally different results after training one extra epoch, we cannot tell which one is correct, especially concerning that there are no ground truth labels for interactions in real world datasets. We will show our method is theoretically locally stable.

For two feed forward neural networks $f$ and $g$ with exactly same architecture, let $\mathcal{G}_f(V,E)$ and $\mathcal{G}_g(V,E)$ be their corresponding weighted graph, respectively.  We denote the measuring function for $f$ and $g$ as $\phi_f$ and $\phi_g$, respectively. For all interaction candidate $\mathcal{I}$ that are both detected in $f$ and $g$ by Algorithm \ref{algo:id}, we denote the interaction strength of $\mathcal{I}$ corresponding to $f$ and $g$ as $\rho_f(\mathcal{I})$ and $\rho_g(\mathcal{I})$ respectively. We propose the following Theorem:

\begin{restatable}[Proof and empirical analysis in Appendix \ref{appendix: lemma stability}]{theo}{theomask}
\label{theomask}
Let $\delta=\max_{e\in E}|\phi_f(e)-\phi_g(e)|$ be the magnitude of perturbation.  For all interaction candidate $\mathcal{I}$ that are both detected in $f$ and $g$ by Algorithm \ref{algo:id}, it satisfies $|\rho_f(\mathcal{I})-\rho_g(\mathcal{I})|\leq C\delta$.
\end{restatable}

Theorem \ref{theomask} only states the stability for interaction candidates that are detected by both $f$ and $g$. 
We note that this is the common situation when the perturbation magnitude is small.
However, there might exists the corner case where there are interaction candidates that are only detected in one network, but not the other. 
We also show the proof that this case only happens if the perturbation magnitude $\delta$ is greater than a threshold in Appendix \ref{appendix: lemma stability}.

\section{Experiments}

Our experiments attempt to answer the following research questions: (\textbf{Q1}) How effective is PID in detecting true interactions (Section \ref{exp: syn})? (\textbf{Q2}) Is the algorithm sensitive to hyperparameters and different architectures (Appendix \ref{appendix: Supplements materials for the synthetic experiments})? (\textbf{Q3}) Is considering these detected interactions beneficial for machine learning models (Section \ref{exp: afe})? (\textbf{Q4}) Can PID detect extremely high-order interactions (Section \ref{exp: image})? We remark the norm $p$ in Algorithm \ref{algo:id} is set to 2 across all experiments, which captures the Euclidean distance of points in persistence diagrams \cite{cohen2010lipschitz}. The other experiment-specific settings are described in respective sections.

\subsection{Pairwise Interaction Detection on Synthetic Data}
\label{exp: syn}

Since there are no ground-truth labels for interactions in real world datasets, to answer \textbf{Q1} and \textbf{Q2}, we utilize ten synthetic datasets that contain a mixture of pairwise interactions and higher-order interactions, as shown in the Appendix~\ref{appendix: experiment setting of syn exp}. For higher-order interactions, we tested their pairwise subsets as in ~\cite{NID,sorokina2008detecting,lou2013accurate}.  All ten datasets and MLP structures are the same as those in~\cite{NID}. The detailed experimental settings can be found in Appendix~\ref{appendix: experiment setting of syn exp}. The pairwise interaction strength of $\{i,j\}$ is obtained by aggregating the strength of all interaction candidates proposed by PID which contain $\{i,j\}$. The layer $l$ in Algorithm 1 is set to the first layer because the neural network naturally separates different interactions in the first hidden layer~\cite{NIT,NID}~(see Figure \ref{fig:stat of x8andx9 in f3} and Figure \ref{fig:weights dist of diff neurons} in Appendix~\ref{appendix:detailed discussion of syn exp}). 

We compared the proposed PID with several strong existing algorithms in the interaction detection literature, including ANOVA~~\cite{fisher1992statistical}, Hierarchical lasso~(HierLasso)~\cite{bien2013lasso}, RuleFit~\cite{friedman2008predictive}, Additive Groves~(AG)~\cite{sorokina2008detecting}, and Neural Interaction Detection~(NID)~\cite{NID}.  Because both PID and NID detect learned interactions from MLPs in a post-hoc way, we apply the NID and PID on the same MLPs for fair comparison. We ran ten trials of AG, NID, and PID on each dataset and removed two trials with the highest and lowest AUC scores. The AUC scores of interaction strength proposed by baseline methods and PID  are shown in Table~\ref{table:result for syn-f}. The heat map of pairwise interaction strength and a detailed analysis about main effects are in~Appendix~\ref{appendix:detailed discussion of syn exp}. Here we provide only the general results.

In general, the AUCs of AG and PID are close, except for $F_5$, $F_6$, and $F_8$, where PID significantly outperforms AG. 
This may be caused by the limitations in the AG's model capacity, which is tree-based \cite{NID}. 
When comparing the AUCs of PID and NID, the AUCs of PID are comparable or better.
We note that PID considers connectivity of the entire NN.
In contrast, NID leverages weights beyond the first hidden layer to obtain the maximum gradient magnitude of the hidden units in the first hidden layer, loosing some information encoded in latter layers in the process. 
Hence, the similar results of NID and PID are likely because the latter layers played lesser roles in this specific setting.
However, we remark PID constantly outperformed NID with various settings, as shown in Appendix E.3, Figure 8, 9, and 10.
To answer \textbf{Q2}, we also compare the result based on MLPs with different architectures~(Appendix \ref{appendix:sentivity of syn exp} Figure~\ref{fig:different architecture default setting}) and regularization strength~(Appendix \ref{appendix:sentivity of syn exp} Figure~\ref{fig:different architecture l1 5e-4}, Figure~\ref{fig:different architecture l1 5e-6}). In general, both NID and PID are insensitive to the architecture of MLPs, and both are sensitive to the regularization strength. 
A possible reason is that the connectivity between hidden units of a trained MLP is significantly influenced by regularization strength. 
We show that the AUCs of PID are better than those of NID under all different settings~(Appendix~\ref{appendix:sentivity of syn exp}).

\begin{table}
  \caption{AUC of pairwise interaction strengths proposed by PID and baselines on the synthetic functions~(Table~\ref{table:syn_f}). ANOVA, HierLasso, and RuleFit are deterministic.}
  \label{table:result for syn-f}
  \centering
\begin{tabular}{c|cccc|cc}
\toprule
   &ANOVA&HierLasso&RuleFit&AG&NID&PID \\
\hline
$F_{1}(x)$ &0.992 &   1.00        &0.754         &\textbf{1$\pm$0.0}    &0.985$\pm$6.3e$-$3     & 0.986$\pm$4.1e$-$3    \\
$F_{2}(x)$ &0.468 &   0.636       &0.698         &\textbf{0.88$\pm$1.4e$-$2 }   &0.776$\pm$4.3e$-$2     &0.804$\pm$5.7e-2     \\
$F_{3}(x)$ &0.657 &   0.556       &0.815         &\textbf{1$\pm$0.0}    &\textbf{1.0$\pm$0.0}     &\textbf{1.0$\pm$0.0}     \\
$F_{4}(x)$ &0.563 &   0.634       &0.689         &\textbf{0.999$\pm$1.4e-3}    &0.916$\pm$6.3e$-$2     &0.935$\pm$3.9e$-$2     \\
$F_{5}(x)$ &0.544 &   0.625        &0.797         &0.67$\pm$5.7e-2    &0.997$\pm$8.9e$-$3     &\textbf{1.0$\pm$0.0}     \\
$F_{6}(x)$ &0.780 &   0.730        &0.811         &0.64$\pm$1.4e-2    &0.999$\pm$3.3e$-$3     &\textbf{1.0$\pm$0.0}     \\
$F_{7}(x)$ &0.726 &   0.571        &0.666         &0.81$\pm$4.9e-2    &0.880$\pm$2.6e$-$2     &\textbf{0.888$\pm$2.8e$-$2}     \\
$F_{8}(x)$ &0.929 &   0.958        &0.946         &0.937$\pm$1.4e-3    &\textbf{1.0$\pm$0.0}   &\textbf{1.0$\pm$0.0}     \\
$F_{9}(x)$ &0.783 &   0.681        &0.584         &0.808$\pm$5.7e-3    &0.968$\pm$2.3e$-$2     &\textbf{0.972$\pm$2.9e$-$2}     \\
$F_{10}(x)$ &0.765 &  0.583         &0.876         &1.0$\pm$0.0    &\textbf{0.989$\pm$3.0e$-$2}     &0.987$\pm$3.5e$-$2     \\
\hline
average &0.721       &  0.698         &0.764         &0.87$\pm$1.4e-2    &0.951$\pm$7.0e$-$2     &\textbf{0.957$\pm$6.2e$-$2}    \\
\hline
\end{tabular}
\end{table}

\subsection{Automatic Feature Engineering}
\label{exp: afe}

\begin{table}[hbt!]
\centering
\caption{Comparing the quality of features automatically generated by interaction detection algorithms. The ``Original'' column shows the results of random forest built without using synthetic features.}
\label{tab:fe_results}
\begin{tabular}{|c|c|c|c|c|}
\hline
Dataset                 & Original & Random & NID & PID \\
\hline
Amazon Employee & 0.8378$\pm$0.0046    &0.7780$\pm$0.0575 & 0.8321$\pm$0.0299  &  \textbf{0.8460$\pm$0.0079}     \\
Higgs Boson     &0.7421$\pm$0.0019 & 0.7421$\pm$0.0192 &      \textbf{0.7422$\pm$0.0017} &  \textbf{0.7422$\pm$0.0017} \\
Creditcard           &0.9555$\pm$0.0390 &0.9579$\pm$0.0377 & 0.9607$\pm$0.0333      &   \textbf{0.9625$\pm$0.0354}  \\
Spambase             &0.9680$\pm$0.0085 &0.9692$\pm$0.0076 &  0.9724$\pm$0.0065    & \textbf{0.9738$\pm$0.0063}\\
Diabetes             &0.8077$\pm$0.0334 &0.8078$\pm$0.0335 &  0.8044$\pm$0.0335     & \textbf{0.8101$\pm$0.0349} \\
\hline          
\end{tabular}
\end{table}

Intricate feature engineering often plays deterministic roles in winning solutions of Kaggle competitions \cite{NIPS2019_9117}. In this regard, interaction detection algorithms are invaluable in that they reveal knowledge about the data. A reasonable question is, can different machine learning models benefit from the knowledge to alleviate the need for hand-crafted feature engineering (\textbf{Q3})? We try to answer it by integrating these detected interactions with the original input features and then check the performance gain of models trained on this augmented data. 


We compare our PID and NID on five real world binary classification datasets. The statistics of these datasets are shown in Appendix~\ref{appendix: Details for automatic feature engineering experiments} Table~\ref{tab:fe_dataset}. Following \cite{tsang2020feature,luo2019autocross}, we explicitly construct synthetic features for each detected interaction candidates and combine these synthetic features with the original feature set. The synthetic feature for interaction $\mathcal{I}$ is the Cartesian product among features in $\mathcal{I}$. The details are described in Appendix~\ref{appendix: Details for automatic feature engineering experiments}. We construct synthetic features for the top ten interactions candidates according to interaction strength. Because of the excellent performance and efficiency of random forest on tabular datasets, similar to~\cite{khurana2018feature, nargesian2017learning}, we choose 
the random forest as our learning algorithm. A more detailed experiment setting can be found in Appendix~\ref{appendix: Details for automatic feature engineering experiments}. In Table \ref{tab:fe_results}, we also report results of the baseline ``Random'', where the ten randomly generated synthetic features are used to combine with the original data and then construct the random forest. The AUCs of the random forest construed with different synthetic features are summarized in Table~\ref{tab:fe_results}.

From Table \ref{tab:fe_results}, we remark that incorporating synthetic features generally boost the performance of the random forest model. Namely, NID outperforms Original in the Higgs Boson, Creditcard, and Spambase datasets, while the proposed PID method outperforms all the compared methods. The statistics of detected interactions by different methods are shown in Appendix \ref{appendix: additional experiment results for autofe exp}. Furthermore, we remark that the feature interactions discovered by PID highly coincide with the top solution of the Amazon Employee Challenge~\footnote[1]{https://www.kaggle.com/c/amazon-employee-access-challenge}(Appendix \ref{appendix: additional experiment results for autofe exp}).


\subsection{High-order Interaction Detection on Image Datasets}
\label{exp: image}


For image data, input features are raw pixels and interactions are patterns that represent visual cues characterizing the object in the image. 
To answer \textbf{Q4}, we apply PID to find out the contributing pattern in a particular image that lead the CNNs to make the prediction. 
In Section~\ref{sec:pd}, the proposed framework is used for detecting global interactions. 
Global (or model-level) interaction means the learned interactions for making predictions on the entire dataset. Specifically, the only input to global interaction detection algorithms is the model to be analyzed, without any information about the position or the scale of the object in an image. 
Local (or instance-level) interaction detection, however, tries to answer what interactions of a data sample lead the model to make a special prediction. 
We remark that global interaction detection is meaningless for image data because it is not invariant to the position or scales of objects. We show how to extend PID to the CNNs and local interaction detection in Appendix \ref{appendix: Extensibility}.

Detecting interactions in a specific image is a more challenging task for the following reasons: 
\begin{wrapfigure}{r}{0.43\textwidth}
    \centering
    \includegraphics[width=0.42\textwidth]{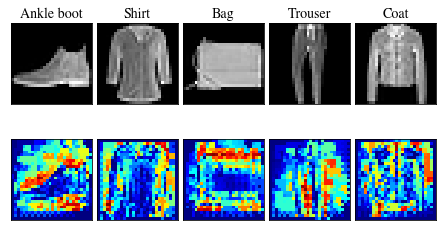}
    \caption{Saliency maps of interaction strength found from applying PID on the CNN trained on FashionMNIST dataset.}
    \label{fig:fashionmnist_pid_res}
\end{wrapfigure}
First, the order of interactions is extremely high in image data;
second, image data is high dimensional by nature. The number of possible interaction candidates grows exponentially with respect to the number of input features, e.g., for a $1\times28\times28$ image from MNIST dataset, the number of  interaction candidates within the search space is $2^{784} \approx 10^{235}$.
We note that interactions in an image is similar to the Superpixels \cite{superpixels}, which is originally proposed for solving the image segmentation task.
However, it is not straight-forward to show detected interaction by considering them as superpixels: 
First, the interaction in an image is a group of pixels that are not necessarily connected; 
second, theoretically, each interaction is associated with the interaction strength, which cannot be shown by simply breaking the image into different segmentation.
To evaluate the detected interactions with better representation of images, 
instead of connecting pixels that are in the same interaction, 
we build the saliency map for each input images by visualizing the importance of raw pixels.
Specifically, the importance of the raw pixel $i$ is obtained by aggregating the interaction strength of each candidate set which contains $i$.

We trained a simple CNN to classify images on the MNIST dataset~\cite{lecun1998gradient} and FashionMNIST dataset~\cite{Fashion-MNIST}. The detailed experiment setting can be found in Appendix~\ref{appendix:Details for local interacion detection on image datasets}. 
Here we only present the saliency maps of FashionMNIST images. The saliency maps of MNIST images are available in Appendix \ref{appendix:Details for local interacion detection on image datasets}. We observe PID is capable of detecting high-order interactions that represent visual cues. From Figure~\ref{fig:fashionmnist_pid_res}, the CNN acquired complex knowledge about the shapes associated with each category. For example, the interpretations of the “Ankle boot” classification show that interaction detection finds the shape of the boot instead of the boot texture. This is indicated by the fact that pixels with higher importance~(warm colors) essentially trace the contour of an ankle boot.

\section{Limitation and Discussion}
In this paper, we extend concepts of the \textit{birth} time and \textit{death} time to the interaction for proposing a new measure that quantifies the interaction strength.
These concepts are originally proposed for identifying true topology features (e.g., connected components and loops). 
Rigorously speaking, the structure of interest for interaction detection is a sub-network that connects a group of input features with output neurons, which is not a well-defined algebraic topology concept. 
Therefore, a lot of theoretical properties of the subject across the filtration is lost. However, to the best of our knowledge, by extending these concepts from Persistent Homology, we propose the first NN specific interaction strength measure with stability guarantee (Theorem 1). 
Furthermore, we derived a topology-motivated algorithm to compute the interaction strength efficiently (Lemma 1).

We note that as NNs contain only 1-simplex, many of these topological properties degenerate to the field of graph theory.
The proposed filtration process is equivalent to building
maximum spanning trees (MSTs) of NNs using the Kruskal algorithm. 
The proposed persistence of feature groups is the gap length between MSTs of two sub-networks.
It would be interesting to consider the theoretical
benefit of our proposed measure from the perspective of graph theory. We leave it as future work.

Also, we want to emphasize that our image experiment in section \ref{exp: image} is exploratory. 
This experiment is designed to illustrate that the proposed PID is capable of detecting extreme high-order interactions in a specific input. Moreover, the saliency map obtained by utilizing PID could also provide visual cues for understanding how CNNs make decisions. 
We note that PID is complementary to most existing explainable-CV works. Especially, the saliency map in section \ref{exp: image} is obtained only from the interaction effects between raw pixels. In contrast, most explainable-CV works (e.g., Grad-CAM \cite{gradcam}) only consider how a specific raw-pixel influence the decision of models, and the interaction effects are ignored by them because most of these works do not access to Hessian Matrix or compute the approximation of Hessian Matrix.

\section{Conclusion}
In this work, we propose a theoretically
well-defined measure for quantifying interaction strength by investigating the topology of neural networks. We show that this measure captures topological information that pertains to learned interactions in neural networks. Based on this measure, we derive the PID algorithm to detect interactions. We also give the theoretical analysis for it and show how to extend our method to local interaction detection. We demonstrate our proposed method has the practical utility of accurately detecting feature interactions without the need to prespecify interaction types or to search an exponential solution space of interaction candidates.

\newpage
\section*{Statement of Broader Impact}

The proposed PID algorithm can be applied in various fields because it provides knowledge about a domain. Any researcher who needs to design experiments might benefit from our proposed algorithm in the sense that it can help researchers formulate hypotheses that could lead to new data collection and experiments. For example, PID can help us discover the combined effects of drugs on human body: 
By utilizing PID on patients' records, we might find using Phenelzine togther with Fluoxetine has a strong interaction effect towards serotonin syndrome.
Thus, PID has great potential in helping the development of new therapies for saving lives.

Also, this project will lead to effective and efficient algorithms for 
finding useful any-order crossing features in an automated way.
Finding useful crossing features is one of the most crucial task in the Recommender Systems.
Engineers and Scientists in E-commerce companies may benefit from our results that our algorithm can alleviate the human effect on finding these useful patterns in the data.


\section*{Acknowledgements}
We would like to sincerely thank everyone who has provided their
generous feedback for this work.
Thank the anonymous reviewers for their thorough comments
and suggestions.
The authors thank the Texas A\&M College of Engineering and Texas A\&M University.

\medskip
\printbibliography

\newpage
\appendix
\appendixpage
\section{The Persistence Interaction Detection Algorithm}
\label{appendix: full PID algo}
\begin{algorithm}
\KwIn{A trained feed-forward neural network, target layer $l$, norm  $p$. }
\KwOut{ranked list of interaction candidates $\{\mathcal{I}_{i}\}$.}
Construct size pair $(\mathcal{G}, \phi)$ and its filtration $\mathcal{G}^{w'_{0}} \subseteq...\subseteq
\mathcal{G}^{w'_{n}}$\\
$\mathcal{K}\leftarrow$ initialize an empty dictionary mapping interaction candidate to persistence

\For{\textrm{$i$=0:n}}{
$\lambda\leftarrow w'_{i}$; \quad $\mathcal{G}^{\lambda}\leftarrow\mathcal{G}^{w'_{i}}$\\
Calculate $\mathbf{M}^{\lambda_{up}}_{(l)}$ and $\mathbf{M}^{\lambda_{down}}_{(l)}$ according to Equation~\eqref{eq:up_down}\\
\For{each row \textbf{m} of $\mathbf{M}^{\lambda_{down}}_{(l)}$ indexed by $r$}{
    \If{all elements in $r^{\mathrm{th}}$ column of $\mathbf{M}^{\lambda_{up}}_{(l)}$ are 0}{
    \Continue\tcp{$r$-th unit in $l$-th layer is not connected with any final output units}
        }
    $\mathcal{I}\leftarrow$ initialize an empty set\;
    \For{$j$=0:$d$-1}{
        \If{$m_{j}$ == 0}{
        \Continue\tcp{$r$-th unit is not connected with feature $j$}
            }
        $d_{\mathcal{I}} \leftarrow \lambda$~\tcp{$\mathcal{I}$ merged with $j$} \;
        $b_{\mathcal{I}\cup j} \leftarrow \lambda$\;
        $\mathcal{K}[\mathcal{I}]\leftarrow\mathcal{K}[\mathcal{I}]+|b_{\mathcal{I}}-d_{\mathcal{I}}|^{p}$\;
        $\mathcal{I} \leftarrow \mathcal{I}\cup j$ \;
        }
    }
}
$\{\mathcal{I}_{i}\}\leftarrow$ interaction candidates in $\mathcal{K}$ sorted by their strengths in descending order.
\caption{The proposed Persistence Interaction Detection~(PID) algorithm}
\label{algo:id}
\end{algorithm}

Our PID framework is presented in Algorithm 1. Besides the $\langle\phi=\lambda\rangle$-connectivity between $\mathcal{I}$ and final outputs, we also consider: 
First, whether $\mathcal{I}$ and a particular neuron $r$ are connected; second, whether the neuron $r$ and final outputs are connected under the threshold $\lambda$. 
Recall that the measuring function $\phi$ is non-decreasing over $\mathcal{G}$~(Definition~\ref{def:filtration}), and the birth time and death time of each interaction candidates can be determined through one pass of all thresholds. 
As shown in Figure \ref{fig:illu of pid}, calculating the interaction strength of $\mathcal{I}$ at neuron $r$ is equivalent to running Algorithm \ref{algo:id} on a neural network whose $l^{\mathrm{th}}$ layer is only composed by neuron $r$.

The time complexity of PID is $\mathcal{O}(Ndp_l)$, where $N$ denotes the total number of weights used as thresholds in the filtration and $p_l$ is the number of neurons at target layer $l$. One possible way to reduce the time complexity is that, we can change the $\mathcal{W'}:=\{|w|/w_{max}|w\in\mathcal{W}\}$ in section \ref{sec:background} to $\mathcal{W'}:=\{|w|/w_{max}|w\in\mathcal{W} \wedge w\geq \eta w_{max}\}$, where $\eta$ is a hyperparameter which controls total number of weights used as thresholds in Algorithm 1. We do not utilize this method to accelerate PID  in all experiments of this paper~(i.e., set $\eta$ as 0).

\section{Proof of Lemma~\ref{lemma:mask}}
\label{appendix:proof for lem 1}
\lemmamask*

We obtain Lemma~\ref{lemma:mask} following from the theoretical analysis in Appendix E of ~\cite{NIT}.

\textit{Proof.}~ If the network has exactly one layer,  $\mathbf{M}^\lambda=(\mathbf{M}_{(1)}^\lambda)^\top$ directly gives the connectivity between input features and output units in the final output layer.


In cases when $\mathbf{M}^\lambda$ has more than one hidden layer, ﬁrst consider the weight connectivity between input features and the second hidden layer. Since a feed-forward neural network is a directed acyclic graph and a hop is a transition from one layer to the next, we can view the connectivity from input features to the second hidden layer as two hops or two applications of an adjacency matrix, $\mathbf{A}$, comprising of $\mathbf{M}_{(2)}^\lambda$ and $\mathbf{M}_{(1)}^\lambda$ as:

\be
\mathbf{A}=
\begin{bmatrix} 
0 & (\mathbf{M}_{(1)}^\lambda)^\top & 0 \\
0 & 0 & (\mathbf{M}_{(2)}^\lambda)^\top\\
0 & 0 & 0 \\
\end{bmatrix}.
\quad\nonumber
\ee
Therefore, the adjacency matrix for two hops is:

\be
\mathbf{A}^2=
\begin{bmatrix} 
0 & 0 & (\mathbf{M}_{(2)}^\lambda)^\top(\mathbf{M}_{(1)}^\lambda)^\top \\
0 & 0 & 0\\
0 & 0 & 0 \\
\end{bmatrix}.
\quad\nonumber
\ee

Since the elements of $\mathbf{A}^2$ are the number of paths between graph vertices in two hops, the non-zero elements of $(\mathbf{M}_{(2)}^\lambda)^\top(\mathbf{M}_{(1)}^\lambda)^\top$ represent the existence of paths from features to the second hidden layer, and the zero elements represent the lack of such paths. We can therefore repeatedly apply hops up to the $L^{th}$ hidden layer, yielding $(\mathbf{M}^{\lambda}_{(L)})^{\top}\cdot(\mathbf{M}^{\lambda}_{(L-1)})^{\top}\cdot\cdot\cdot(\mathbf{M}^{\lambda}_{(1)})^{\top}$ to represent the zero and non-zero paths from input features to the neurons in the $L^{th}$ layer. Thus, if all elements in $[\mathbf{m}^{\lambda}_{r}]^{\mathcal{I}}$ are non-zero and all other elements in $[\mathbf{m}^{\lambda}_{r}]^{\{0,...,d-1\}\backslash \mathcal{I}}$ are zero, $\mathcal{I}$ and unit $r$ are $\langle\phi=\lambda\rangle-connected$ by Definition~\ref{def:connected_path}. 

\qedwhite
 
\section{Proof of Theorem~\ref{theomask}}
\label{appendix: lemma stability} 

In this subsection, we will prove Theorem~\ref{theomask} and evaluate it empirically. 
We first give the stability lemma for connected components and then utilize it to derive Theorem \ref{theomask}. 

\begin{definition}[Hausdorff distance]
For points $p=(p_{1}, p_{2})$ and $q=(q_{1}, q_{2})$ in $\mathbb{R}^{2}$, 
let $\|p-q\|_{\infty}$ be the maximum of $|p_{1} - q_{1}|$ and $|p_{2} - q_{2}|$. 
Let $\|f-g\|_{\infty}=\mathrm{sup}_{x}|f(x)-g(x)|$. 
Let X and Y be multisets of points. 
The Hausdorff distance is defined as
\be
d_{H}(X, Y) = \mathrm{max}\{ \mathrm{sup}_{x\in X} \mathrm{inf}_{y\in Y}\|x-y\|_{\infty}, 
\mathrm{sup}_{y\in Y} \mathrm{inf}_{x\in X}\|x-y\|_{\infty}\}, \nonumber
\ee
\end{definition}

For two feed forward neural networks $f$ and $g$ with the exact same architecture, let $g$ be a neural network that is obtained by perturbing the weights of $f$. The corresponding size pairs ($\mathcal{G}_f,\phi_f$) and ($\mathcal{G}_g,\phi_g$) are constructed following instructions in Section \ref{sec:background}. Let $\delta=\max_{e\in E}|\phi_f(e)-\phi_g(e)|$ be the magnitude of the perturbation, i.e., $\|\phi_f-\phi_g\|_{\infty}=\delta$. Persistence diagrams of ($\mathcal{G}_f,\phi_f$) and ($\mathcal{G}_g,\phi_g$) are denoted as $\mathcal{D}[(\mathcal{G}_f,\phi_f)]$ and $\mathcal{D}[(\mathcal{G}_g,\phi_g)]$, respectively. We note that $\phi_f$ and $\phi_g$ are piecewise linear functions on simplicial complexes, where a simplicial complex is a high-dimensional generalization of a graph in topological space. Piecewise linear functions satisfy the following Lemma:

\begin{lemma}[Proof in~\cite{cohen2007stability}]
\label{lem:stability}
$d_{H}(\mathcal{D}[(\mathcal{G}_f,\phi_f)],\mathcal{D}[(\mathcal{G}_g,\phi_g)])\leq \delta$.
\end{lemma}

When weights in the networks are perturbed, 
the birth time and death time of connected components are also changed. Lemma~\ref{lem:stability} shows that the Hausdorff distance between the persistence diagrams is bounded by the magnitude of the perturbation, i.e., for the set of all connected components $\mathcal{J}$, suppose its birth time~$b_{\mathcal{J}}$ and death time $d_{\mathcal{J}}$ changes to $b'_{\mathcal{J}}$ and $d'_{\mathcal{J}}$, then $\mathrm{max}(|b_{\mathcal{J}}-b'_{\mathcal{J}}|, |d_{\mathcal{J}}-d'_{\mathcal{J}}|)\leq \delta$.

For any interaction candidate $\mathcal{I}$ that are detected in both $f$ and $g$ by Algorithm 1, we denote the birth time of $\mathcal{I}$ in $f$ and $g$ as $b_{\mathcal{I}}$ and $b'_{\mathcal{I}}$, respectively. Similarly, we use $d_{\mathcal{I}}$ and $d'_{\mathcal{I}}$ for the death time of $\mathcal{I}$ in $f$ and $g$, respectively.
Suppose the connected component $\mathcal{J}$ and the connected component $\mathcal{J'}$ cause the birth of interaction $\mathcal{I}$ in $f$ and $g$, respectively. 
We have the following corollary:

\begin{corol}
\label{corol:distance of birth time of interaction}
$|b_\mathcal{I} - b'_\mathcal{I}| \leq 3\delta$.
\end{corol}

\textit{Proof.}~~From Definition \ref{def:connected_path}, we have $|b_\mathcal{I} - b'_\mathcal{I}| = |b_{\mathcal{J}}-b_{\mathcal{J'}}|=|\min_{e\in\mathcal{J}}\phi_f(e)-\min_{e\in\mathcal{J'}}\phi_g(e)|$. If $\mathcal{J}$ and $\mathcal{J'}$ are composed of identical set of edges, then we can directly prove Corollary \ref{corol:distance of birth time of interaction} following Lemma \ref{lem:stability}. If $\mathcal{J}$ and $\mathcal{J'}$ contain different edges, without loss of generality, let $\mathcal{J'}\backslash\mathcal{J}\neq \emptyset$. $\forall e, \phi_f(e)\leq \min_{e'\in\mathcal{J} }\phi_f(e') - 2 \delta$, we have the following inequality:
\be
\label{eq: eqe}
\phi_g(e)\leq \min_{e'\in\mathcal{J} }\phi_g(e').
\ee
Inequality (\ref{eq: eqe}) follows from the fact that $\forall e, |\phi_f(e) - \phi_g(e)|\leq \delta$,
which implies that, $\forall e, \phi_f(e)\leq \min_{e'\in\mathcal{J} }\phi_f(e') - 2 \delta$, $e$ has to wait for all edges in $\mathcal{J}$ to be added to the filtration before being added itself.
Namely, $\mathcal{I}$ is born before the threshold arrives at $\phi_g(e)$ and, consequently, $e\notin \mathcal{J'}$. Thus, $\forall e, e\in \mathcal{J'} \backslash \mathcal{J}$, $e$ satisfies $\phi_f(e)\geq \min_{e'\in\mathcal{J} }\phi_f(e') - 2 \delta$. Following this fact, we have

\be
\min_{e\in\mathcal{J'}}\phi_g(e) 
    &=& \min\{\min_{e\in\mathcal{J'}\cup\mathcal{J}}\phi_g(e),
        \min_{e\in\mathcal{J'}\backslash\mathcal{J} }\phi_g(e)\} \nonumber \\
    \label{eq: 1 inequ}
    &\geq& \min\{\min_{e\in\mathcal{J'}\cup\mathcal{J}}\phi_f(e) - \delta,
        \min_{e\in\mathcal{J'}\backslash\mathcal{J} }\phi_f(e) - \delta\} \\
    \label{eq: 2 inequ}
    &\geq& \min\{\min_{e\in\mathcal{J'}\cup\mathcal{J}}\phi_f(e) - \delta,
        \min_{e\in\mathcal{J} }\phi_f(e) - 3 \delta\} \\
    &\geq& \min\{\min_{e\in\mathcal{J}}\phi_f(e) - \delta,
        \min_{e\in\mathcal{J} }\phi_f(e) - 3 \delta\} \nonumber \\
    \label{eq: 3rd eq }
    &=& \min_{e\in\mathcal{J} }\phi_f(e) - 3 \delta.\
\ee

Inequality (\ref{eq: 1 inequ}) follows from $\forall e\in E, |\phi_f(e)-\phi_g(e)|\leq\delta$. 
The inequality (\ref{eq: 2 inequ}) follows from the fact that $\forall e\in \mathcal{J'}\backslash\mathcal{J}, \phi_f(e)\geq \min_{e'\in\mathcal{J} }\phi_f(e') - 2 \delta$. By equation (\ref{eq: 3rd eq }), we have $b_{\mathcal{J}}-b_{\mathcal{J'}}\leq 3\delta$. 

 By exchanging $f$ with $g$, we have $b_{\mathcal{J'}}-b_{\mathcal{J}}\leq 3\delta$. Combining them together finishes the proof. 

 \qedwhite

It is trivial to show that Corollary \ref{corol:distance of birth time of interaction} can be extended to the death time, i.e., we also have $|d_\mathcal{I} - d'_\mathcal{I}| \leq 3\delta$.

After proving Corollary~\ref{corol:distance of birth time of interaction}, we return to prove the theorem. 
\theomask*

\textit{Proof.}~~
In Algorithm \ref{algo:id}, interaction candidates are generated at each neuron $r$ of a particular layer $l$. 
As shown in Figure \ref{fig:illu of pid}, calculating the interaction strength of $\mathcal{I}$ at neuron $r$ is equivalent to running Algorithm \ref{algo:id} on a neural network whose $l^{\mathrm{th}}$ layer is only composed by neuron $r$. 
Thus Corollary \ref{corol:distance of birth time of interaction} also holds for interaction candidate $\mathcal{I}$ generated at each neuron. 
We use $\mathrm{per}^{(r)}_f(\mathcal{I})$, $b^{(r)}(\mathcal{I})$, and $d^{(r)}(\mathcal{I})$ to represent the persistence, the birth time, and the death time of $\mathcal{I}$ generated at neuron $r$ corresponding to $f$, respectively. 
Similarly, for $g$, the persistence, the birth time, and the death time of $\mathcal{I}$ generated at neuron $r$ are denoted as $\mathrm{per}^{(r)}_g(\mathcal{I})$, $b'^{(r)}(\mathcal{I})$, and  $d'^{(r)}(\mathcal{I})$ , respectively.

\be
\mathrm{per}^{(r)}_f(\mathcal{I}) 
    &=& |b^{(r)}_\mathcal{I}-d^{(r)}_\mathcal{I}| \nonumber\\
    &=& |b^{(r)}_\mathcal{I} - b'^{(r)}_\mathcal{I} + b'^{(r)}_\mathcal{I} - d'^{(r)}_\mathcal{I} + d'^{(r)}_\mathcal{I} - d^{(r)}_\mathcal{I}| \nonumber\\
    &\leq& \mathrm{per}^{(r)}_g(\mathcal{I}) + 6 \delta, \nonumber
\ee


By exchanging $f$ with $g$, we have $\mathrm{per}^{(r)}_g(\mathcal{I})\leq 6\delta + \mathrm{per}^{(r)}_f(\mathcal{I})$. 
Combining them together, we have $|\mathrm{per}^{(r)}_f(\mathcal{I})-\mathrm{per}^{(r)}_g(\mathcal{I})|\leq 6\delta$. 
Then it follows
\be
|\rho_f(\mathcal{I})-\rho_g(\mathcal{I})| &=& |\sum_{r\in l^{th} \mathrm{layer}} [\mathrm{per}^{(r)}_f(\mathcal{I})]^p - [\mathrm{per}^{(r)}_g(\mathcal{I})]^p|\nonumber\\
&\leq& p|\sum_{r\in l^{th} \mathrm{layer}}[\mathrm{per}^{(r)}_f(\mathcal{I}) - \mathrm{per}^{(r)}_g(\mathcal{I})]\max\{\mathrm{per}^{(r)}_f(\mathcal{I}),\mathrm{per}^{(r)}_f(\mathcal{I})\}^{p-1}| \nonumber\\
\label{eq: 4 eq}
&\leq& 6p N_{l} \delta.
\ee
Where $N_l$ is the number of units in layer $l$. The inequality (\ref{eq: 4 eq}) follows from the fact that $\max\{\mathrm{per}^{(r)}_f(\mathcal{I}),\mathrm{per}^{(r)}_f(\mathcal{I})\}\leq 1$. 

\qedwhite

Beyond Theorem \ref{theomask}, there exists the corner case that there are interaction candidates only detected in one neural network, but not the other.
We will show that this corner case only happens if $\delta$ is greater than a threshold.

Let $[d]:=\{0,\cdots,d-1\}$ be the input feature set. Without loss of generality, suppose interaction candidate $\mathcal{I}\subset [d]$ only born in $f$, but not in $g$; and the connected component $\mathcal{J}$ cause the birth of $\mathcal{I}$ in $f$.
Let $g$ be a neural network that is obtained by perturbing the weights of $f$.
According to Definition \ref{def:connected_path}, if $\mathcal{I}$ only born in $f$, it means that there exists some edges corresponding to the connection between input features and hidden units in the first layer, which satisfy the following:
\be
\exists e'\in [d]\backslash\mathcal{I}, \nonumber\\
s.t.~~~\phi_f(e') \geq \min_{e\in\mathcal{J}} \phi_f(e) - 2\delta
\ee
The above inequality follows from the fact that 
$\forall e, |\phi_f(e)-\phi_g(e)|\leq\delta$,
which implies that, $\forall e, \phi_f(e)\leq \min_{e'\in\mathcal{J} }\phi_f(e') - 2 \delta$, $e$ has to wait for all edges in $\mathcal{J}$ to be added to the filtration before being added itself. 
Therefore, if $\mathcal{I}$ does not born in $g$,
there must $\exists e'\in [d]\backslash\mathcal{I}$ such that 
$\phi_f(e') \geq \min_{e\in\mathcal{J}} \phi_f(e) - 2\delta$.
In conclusion, if $\mathcal{I}$ only detected in $f$, 
the perturbation magnitude $\delta$ must satisfy:
\be
\label{eq: perb mag upper bound}
\delta \geq \frac{\min_{e\in\mathcal{J}}\phi_f(e) - \max_{e\in[d]\backslash\mathcal{I}} \phi_f(e)}{2}
\ee

\begin{table}[hbt!]
\centering
\caption{Perturbation results.}
\label{tab:perturb results}
\begin{tabular}{|c|c|} 
\hline
$\delta$                    & $|\rho_f(\mathcal{I})-\rho_g(\mathcal{I})|$   \\
\hline
0.001 & 0.0935                   \\
0.01                        & 0.1645                                        \\
0.1                         & 0.1990                                        \\
1                           & 0.3321                                        \\
\hline
\end{tabular}
\end{table}

Here we randomly perturb the weights of an MLP trained on synthetic dataset $F_1$, which has architecture of 64-32-16 first-to-last hidden layer sizes. The layer $l$ in Algorithm \ref{algo:id} is set to the first layer, and the norm $p$ in Algorithm \ref{algo:id} is set to 2. The results are shown in Table~\ref{tab:perturb results}.

\section{Extensibility}
\label{appendix: Extensibility}
In this section, first, we show how to extend PID to CNNs. Second, we introduce how to extend our method to local interaction detection.

Let $\mathbf{H}\in\mathbb{R}^{height\times width}$ be a convolution kernel and $\mathbf{X}\in\mathbb{R}^{H\times W}$ be a tensor. Let $*$ refer to the convolution operation. Suppose the height and width of the $\mathbf{H}*\mathbf{X}$ are $H_{\mathrm{out}}$ and $W_{\mathrm{out}}$, respectively. We define $\mathcal{H}\in\mathbb{R}^{H_{\mathrm{out}}\times W_{\mathrm{out}} \times H \times W}$ as the corresponding four dimensional convolution tensor such that:
\be
\mathcal{H}(i,j,i:i+height,j:j+width)=\mathbf{H}, \nonumber
\ee
for $\forall i\in [0, H_{\mathrm{out}}), \forall j\in [0, W_{\mathrm{out}})$. Then we have the following equation:
\be
\label{eq:conv_tensor}
\mathbf{H}*\mathbf{X}=\mathcal{H}\otimes\mathbf{X},
\ee
where $\mathcal{H}\otimes\mathbf{X}$ is the tensor product such that $[\mathcal{H}\otimes\mathbf{X}]_{i,j}=\sum_{k=0}^{H}\sum_{l=0}^{W} \mathcal{H}_{i,j,k,l}X_{k,l}$. Generally, for 
$\mathbf{H'}\in\mathbb{R}^{C_{\mathrm{out}}\times C_{\mathrm{in}}\times height\times width}$ and $\mathbf{X}'\in\mathbb{R}^{C_{\mathrm{in}} \times H\times W}$, we can convert the convolution between $\mathbf{H'}$ and $\mathbf{X}'$ into $\mathbf{H'}*\mathbf{X'}=\mathcal{H''}\mathbf{X''}$~\cite{kong2017take} using equation \eqref{eq:conv_tensor}, where $\mathcal{H''}\in\mathbb{R}^{C_{\mathrm{out}}H_{\mathrm{out}}W_{\mathrm{out}}\times C_{\mathrm{in}}HW}$ is the flattened matrix of $\mathcal{H'}$ with multi-channels, and $\mathbf{X''}\in\mathbb{R}^{C_{in}HW}$ is the flattened vector of $\mathbf{X'}$. Then we can build the filtration of CNNs just as MLPs.

Given a data point $\mathbf{x}\in\mathbb{R}^{d}$, the feed forward neural network $f$ with \textrm{ReLU} activation function is a linear model in a region surrounding $\mathbf{x}$:
\be
\label{eq:lid}
f(\mathbf{x})=\mathbf{W}^{(L)\top}_{\mathbf{x}}\cdots \mathbf{W}^{(1)\top}_{\mathbf{x}} \mathbf{x},
\ee
where $\mathbf{W}^{(L)}_{\mathbf{x}}$ is the equivalent weight matrix which combines the resulted activation pattern with $\mathbf{W}^{(L)}$, e.g., $\mathrm{ReLU}(\mathbf{W}^{(1)\top}\mathbf{x})=\mathbf{W}^{(1)\top}_{\mathbf{x}} \mathbf{x}$, where $\mathbf{W}^{(1)}_{\mathbf{x}}$ is modified from $\mathbf{W}^{(1)}$ by setting the columns, whose corresponding activation patterns are 0, to be all zero vectors. We denote the output value of the $i^{th}$ neuron in the $l^{th}$ layer, before activation, as $z^{l}_{i}$. From equation (\ref{eq:lid}), for local interaction detection, 
the measuring function $\phi$ can be revised to 
\be
\phi((v_{l-1,i},v_{l,j}))=\frac{{}|W^{(l)}_{i,j}|\mathrm{ReLU}(z^{l-1}_{i})}{\Phi},
\ee
where $\Phi = \max_{i,j,l}|W^{(l)}_{i,j}|\mathrm{ReLU}(z^{l-1}_{i})$.
\section{Supplemental Material for the Synthetic Data Experiments}
\label{appendix: Supplements materials for the synthetic experiments}
\subsection{Experiment Setting}
\label{appendix: experiment setting of syn exp}
\begin{table}[hbt!]
  \caption{Test suite of data-generating functions.}
  \label{table:syn_f}
  \centering
  \begin{tabular}{l|l}
    \toprule
    $F_{1}(x)$ & $\pi^{x_{0}x_{1}}\sqrt{2x_{2}}-\mathrm{sin^{-1}}(x_{3})+\mathrm{log}(x_{2}+x_{4})-
    \frac{x_{8}}{x_{9}}\sqrt{\frac{x_{6}}{x_{7}}}-x_{1}x_{6}$      \\
    \hline
    $F_{2}(x)$     & $\pi^{x_{0}x_{1}}\sqrt{2|x_{2}|}-\mathrm{sin^{-1}}(0.5x_{3})+\mathrm{log}(|x_{2}+x_{4}|+1)+ \frac{x_{8}}{1+|x_{9}|}\sqrt{\frac{x_{6}}{1+|x_{7}|}}-x_{1}x_{6}$       \\
    \hline
    $F_{3}(x)$     & $e^{|x_{0}-x_{1}|}+|x_{1}x_{2}|-x_{2}^{2|x_{3}|}+\mathrm{log}(x_{3}^2+x_{4}^2+x_{6}^2+x_{7}^2)+x_{8}+\frac{1}{1+x_{9}^2}$        \\
    \hline
    $F_{4}(x)$     & $e^{|x_{0}-x_{1}|}+|x_{1}x_{2}|-x_{2}^{2|x_{3}|}+\mathrm{log}(x_{3}^2+x_{4}^2+x_{6}^2+x_{7}^2)+x_{8}+\frac{1}{1+x_{9}^2}+x_{0}^{2}x_{3}^{2}$       \\
    \hline
    $F_{5}(x)$     & $\frac{1}{1+x_{0}^{2}+x_{1}^{2}+x_{2}^{2}}+\sqrt{e^{x_{3}+x_{4}}}+|x_{5}+x_{6}|+x_{7}x_{8}x_{9}$      \\    
    \hline
    $F_{6}(x)$     & $e^{|x_{0}x_{1}+1|}-e^{|x_{2}+x_{3}|+1}+\mathrm{cos}(x_{4}+x_{5}-x_{7})+\sqrt{x_{7}^{2}+x_{8}^{2}+x_{9}^{2}}$        \\
    \hline
    $F_{7}(x)$     & $(\mathrm{tan^{-1}}x_{0}+\mathrm{tan^{-1}}x_{1})^{2}+\mathrm{max}(x_{2}x_{3}+x_{5},0)-\frac{1}{1+(x_{3}x_{4}x_{5}x_{6}x_{7})^{2}}+(\frac{|x_{6}|}{1+|x_{8}|})^{5}+\sum_{i=0}^{9}x_{i}$\\
    \hline
    $F_{8}(x)$     & $x_{0}x_{1}+2^{x_{2}+x{4}+x_{5}}+2^{x_{2}+x_{3}+x_{4}+x_{6}}+\mathrm{sin}(x_{6}\mathrm{sin}(x_{7}+x_{8}))+\mathrm{cos^{-1}}(0.9x_{9})$ \\
    \hline
    $F_{9}(x)$     & $\mathrm{tanh}(x_{0}x_{1}+x_{2}x_{3})\sqrt{|x_{4}|}+e^{x_{4}+x_{5}}+\mathrm{log}(x_{5}^{2}x_{6}^{2}x_{7}^{2}+1)+x_{8}x_{9}+\frac{1}{1+|x_{9}|}$\\
    \hline
    $F_{10}(x)$     & $\mathrm{sinh}(x_{1}+x_{2})+\mathrm{cos^{-1}(tanh(}x_{2}+x_{4}+x_{6}))+\mathrm{cos}(x_{3}+x_{4})+\mathrm{sec}(x_{6}x_{8})$\\
    \bottomrule
  \end{tabular}
\end{table}

The synthetic datasets in section~\ref{exp: syn} are shown in Table~\ref{table:syn_f}. $F_1$ is a commonly used function in interaction detection literature~\cite{hooker2004discovering, lou2013accurate, sorokina2008detecting}. All features were uniformly distributed between -1 and 1 except in $F_1$, where we used the same variable ranges as those reported in ~\cite{lou2013accurate}. In all synthetic
experiments, we evenly split train set, validation set and test set on 30k data points. All networks consisted of four hidden layers with first-to-last layer sizes of: 140, 100, 60, and 20 units.  All networks employed ReLU activation and were trained using Adam optimizer with a $5e-3$ learning rate cross all ten datasets. The L1 regularization strength was set to $5e-5$. The early stopping round was set to 100 to prevent overfitting. The mean square error of all trained MLPs are less than $3e-3$ on test data.

\subsection{Detailed Analysis}
\label{appendix:detailed discussion of syn exp}

\begin{figure}[hbt!]
    \centering
    \includegraphics[scale=0.25]{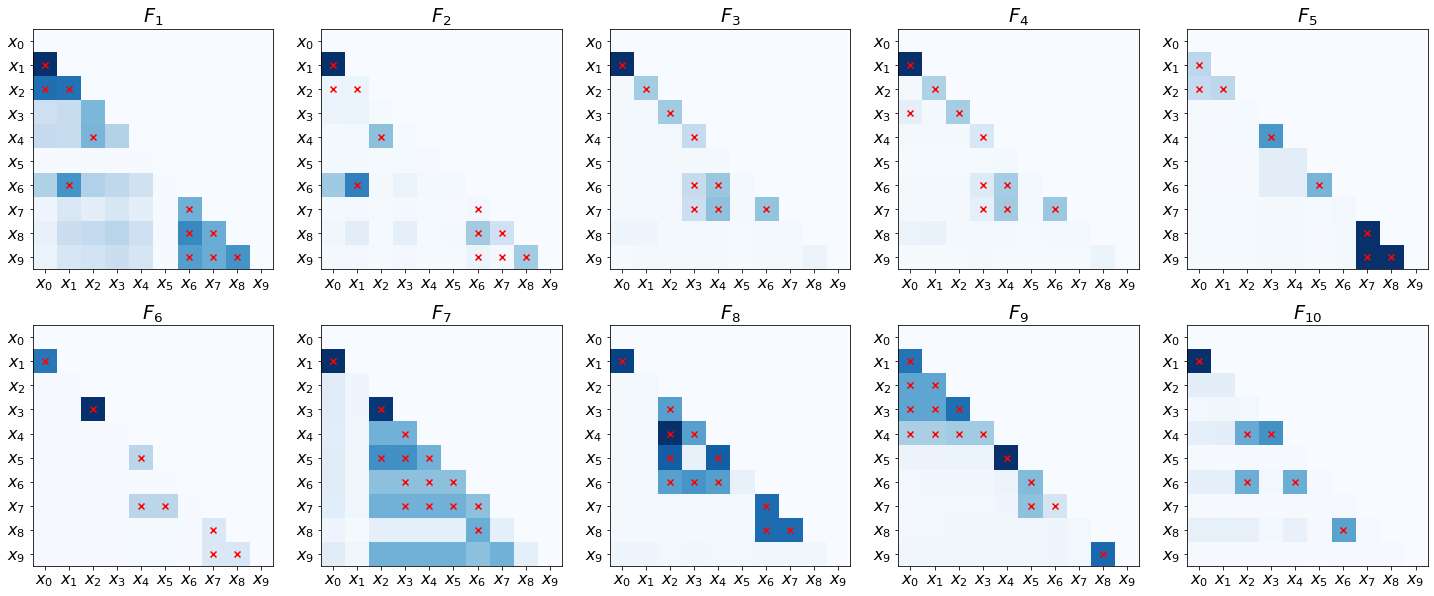}
    \caption{Heat maps of pairwise interaction strengths proposed by our PID corresponding to Table~\ref{table:result for syn-f}. Cross-marks indicate ground truth interactions.}
    \label{fig:heat map of syn}
\end{figure}

Main effects describe the univariate influences of features on outcomes~\cite{friedman2008predictive}, e.g., $\sin^{-1}(0.5x_3)$ in the synthetic dataset $F_2$. Main effects might entangle with true interactions, resulting in spurious interactions. For example, in $F_2$, $\{0,1,2\}$ is true interaction and $\{0,1,2,3\}$ is a spurious interaction, which is an entanglement between true interaction $\{0,1,2\}$ and main effect $\{3\}$. Handling main effects is an important problem in interaction detection~\cite{lim2015learning,bien2013lasso,kong2017interaction}. We remark that in synthetic experiments, higher AUCs indicates the interaction detection algorithms can more thoroughly disentangle main effects from true interactions.

In Figure~\ref{fig:heat map of syn}, heat maps of synthetic functions show the relative strengths of all possible pairwise
interactions proposed by PID, and the ground truth is indicated by red cross-marks. In general, the interaction strengths are higher at the cross-marks. Although most of the synthetic functions contain main effects, from Figure~\ref{fig:heat map of syn} and Table~\ref{table:result for syn-f}, the influence of main effects is limited: only the AUCs of $F_2$ and $F_7$ are under 0.9. We hypothesize that if a overparameterized neural network is trained with proper regularization, the neural network will push the modeling of main effect to a small portion of neurons at the first layer.

To confirm our hypothesis, here we analyze the MLP trained on synthetic dataset $F_3$. For $F_3$, main effects are $x_8$ and $\frac{1}{1+x^2_{9}}$. Let $\mathbf{W}^{(1)}\in\mathbb{R}^{d\times p_{1}}$ be the weight matrix of the first layer. The weights corresponding to input feature $r$ are the $r^{th}$ row of $\mathbf{W}^{(1)}$, which is denoted as $\mathbf{W}^{(1)}_{r,:}$. For convince, we mark different neurons at the first layer by their indices. In Figure~\ref{fig:stat of x8andx9 in f3}, we show the statistics of magnitudes of $\mathbf{W}^{(1)}_{8,:}$ and $\mathbf{W}^{(1)}_{9,:}$ of an MLP trained on synthetic dataset $F_3$. In general, only a few neurons have large weights connecting to $x_8$ and $x_9$, which are corresponding to the peaks in Figure~\ref{fig:stat of x8andx9 in f3}. We plot the weights of all input features  to these neurons in Figure~\ref{fig:weights dist of diff neurons}. To be specific, given a representative neuron $c$, we plot the weight statistics of input features to that neuron, which is denoted as $\mathbf{W}^{(1)}_{:,c}$. For $\mathbf{W}^{(1)}_{9,:}$, two peaks in Figure \ref{fig:stat of x8andx9 in f3} have identical patterns. Here we only show statistics for one of them. For $\mathbf{W}^{(1)}_{8,:}$, we show weights statistics of all input features to neuron 36; For $\mathbf{W}^{(1)}_{9,:}$,  we show weights statistics of all input features to neuron 53. This result is consistent with our hypothesis: neural networks will naturally separate different interactions in the first hidden layer.

\begin{figure}[hbt!]
\centering
\begin{subfigure}[h]{0.4\linewidth}
\centering
\includegraphics[width=\linewidth]{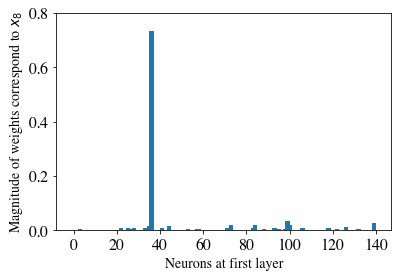}
\end{subfigure}\hspace{0.05\textwidth}
\begin{subfigure}[h]{0.4\linewidth}
\centering
\includegraphics[width=\linewidth]{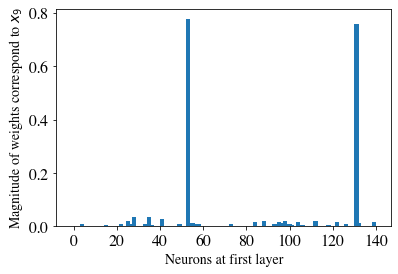}
\end{subfigure}
\caption{Statistics of the magnitudes of $\mathbf{W}^{(1)}_{8,:}$ and $\mathbf{W}^{(1)}_{9,:}$~(the MLP is trained on $F_3$).}
\label{fig:stat of x8andx9 in f3}
\end{figure}

\begin{figure}[hbt!]
\centering
\begin{subfigure}[h]{0.4\linewidth}
\centering
\includegraphics[width=\linewidth]{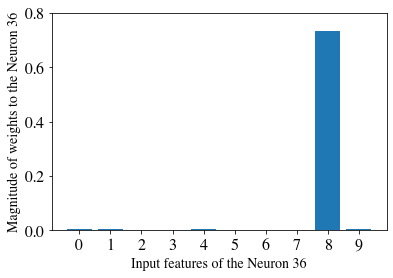}
\end{subfigure}\hspace{0.05\textwidth}
\begin{subfigure}[h]{0.4\linewidth}
\centering
\includegraphics[width=\linewidth]{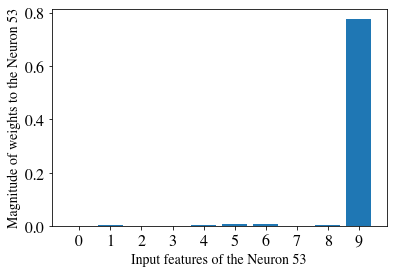}
\end{subfigure}
\caption{The magnitude of weights corresponding to different input features at the selected representative neurons in the first layer~(these neurons are corresponding to the peaks in Figure \ref{fig:stat of x8andx9 in f3}).}
\label{fig:weights dist of diff neurons}
\end{figure}

\subsection{Sensitivity to the Architecture and Regularization Strength}
\label{appendix:sentivity of syn exp}
We try to analyze the sensitivity of interaction detection algorithms to the architecture of MLPs. In Figure~\ref{fig:different architecture default setting}, 64 represents an MLP with first-to-last layer sizes of 64-32-16; 128 represents an MLP with the 128-64-32 architecture; 140 represents an MLP with the 140-100-60-20 architecture; and 256 represents an MLP with the 256-128-64 architecture. The training hyperparameters of these MLPs are identical to those reported in Appednix \ref{appendix: experiment setting of syn exp}. We ran ten trials of NID and PID on each dataset and removed two trials with the highest and the lowest AUC scores.
The mean square errors of all MLP models used for detecting interactions are less than 3e$-$3 on test data.

\begin{figure}[hbt!]
\centering
\begin{subfigure}[h]{0.4\linewidth}
\centering
\includegraphics[width=\linewidth]{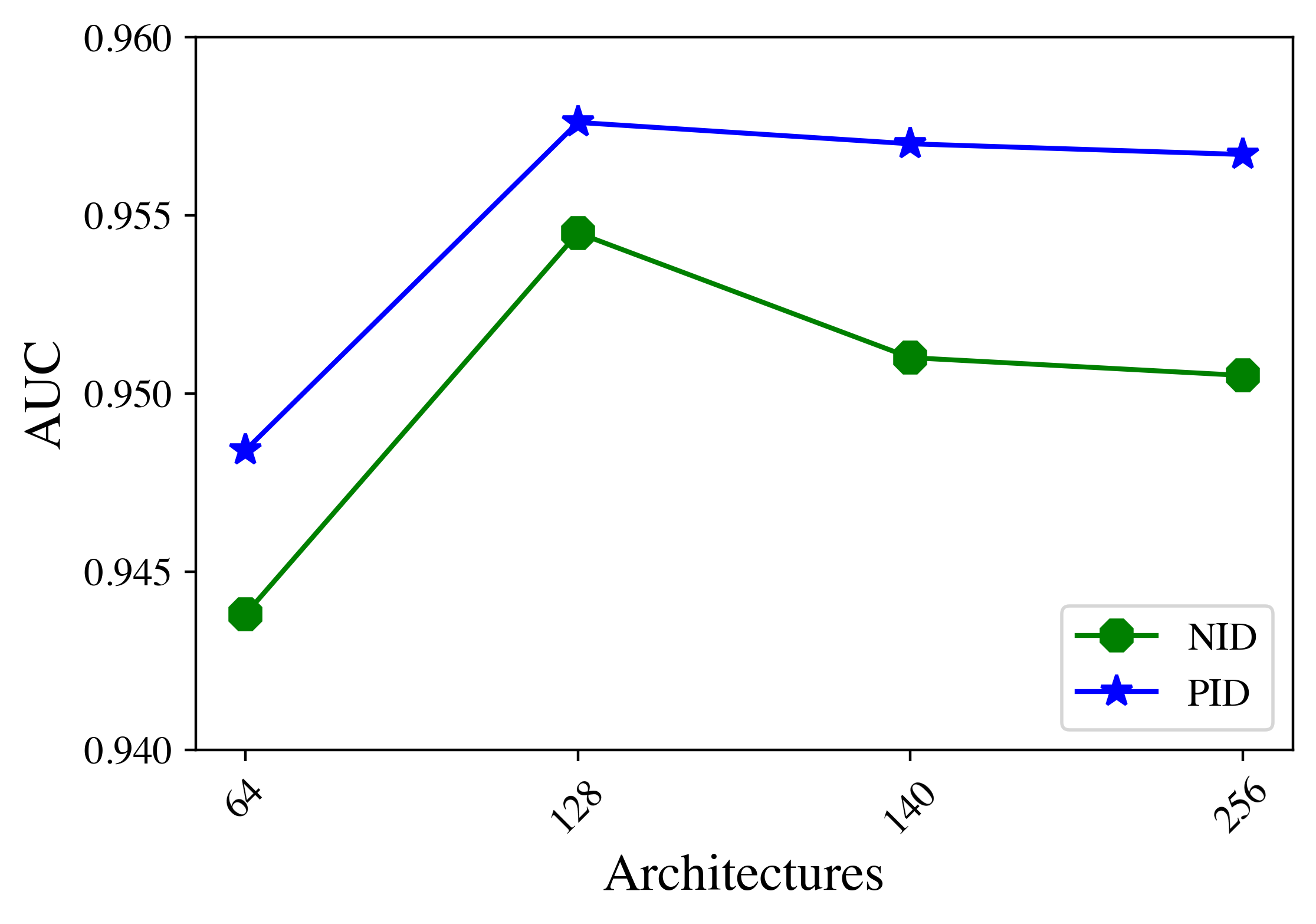}
\caption{Average AUCs of pairwise interaction detected by NID and PID using MLPs with different architectures.}
\end{subfigure}\hspace{0.05\textwidth}
\begin{subfigure}[h]{0.4\linewidth}
\centering
\includegraphics[width=\linewidth]{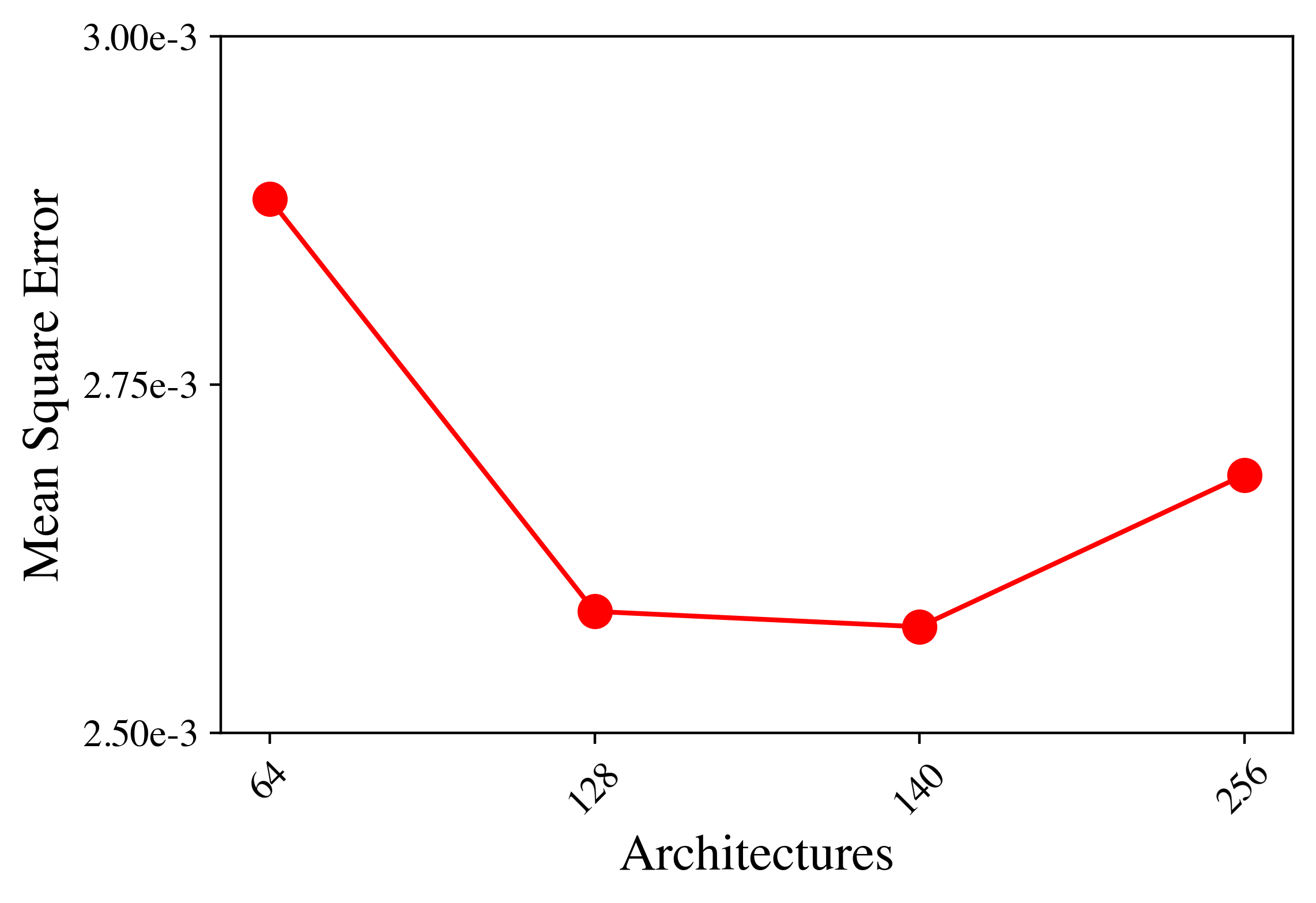}
\caption{Average Mean Square Error~(MSE) of MLPs with different architectures on test data.}
\end{subfigure}
\caption{The sensitivity analysis of interaction detection algorithms to the architecture of MLPs ~(L1 is set to $5e-5$).}
\label{fig:different architecture default setting}
\end{figure}

\begin{figure}[hbt!]
\centering
\begin{subfigure}[h]{0.4\linewidth}
\centering
\includegraphics[width=\linewidth]{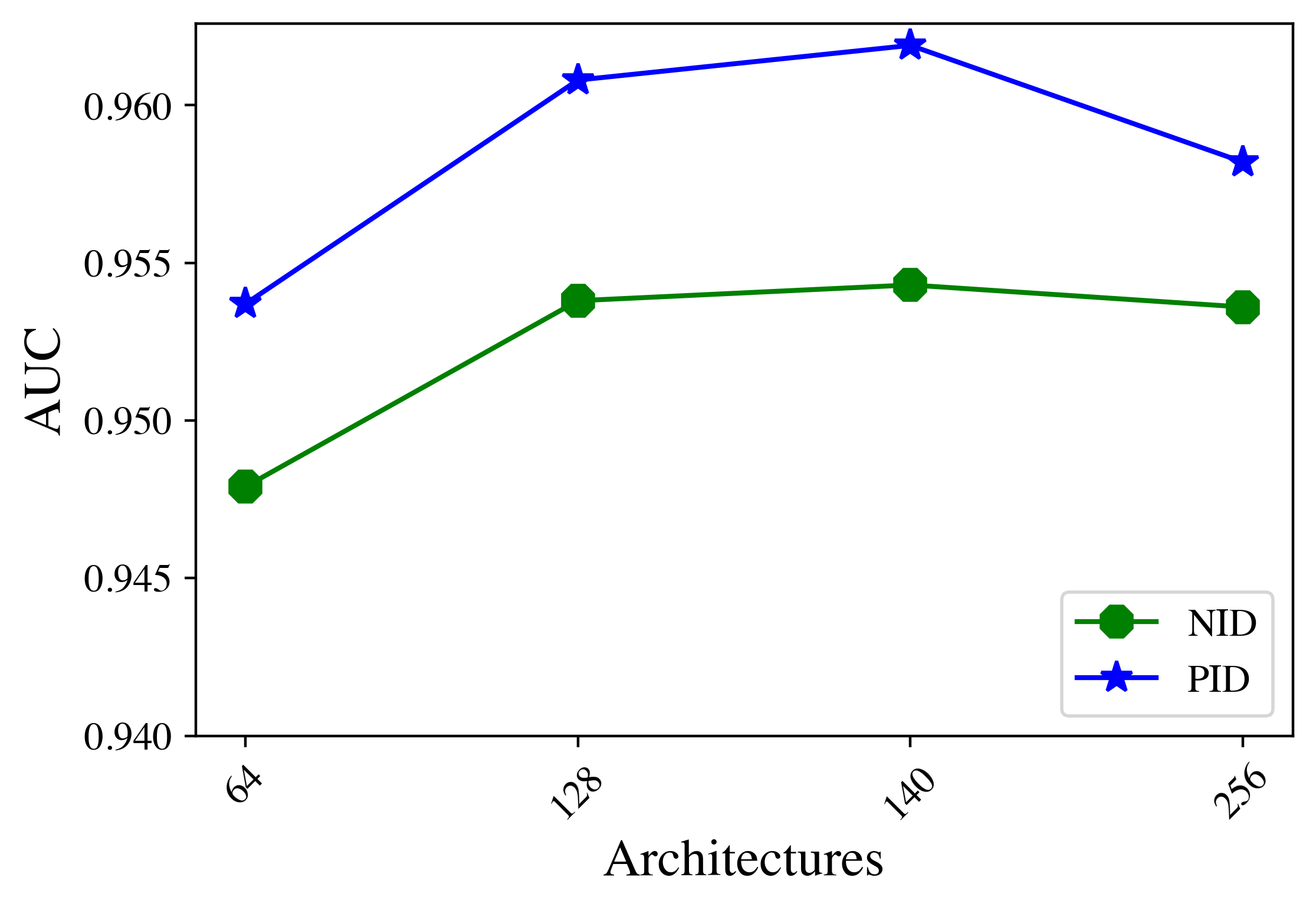}
\caption{Average AUCs of pairwise interaction detected by NID and PID using MLPs with different architectures.}
\end{subfigure}\hspace{0.05\textwidth}
\begin{subfigure}[h]{0.4\linewidth}
\centering
\includegraphics[width=\linewidth]{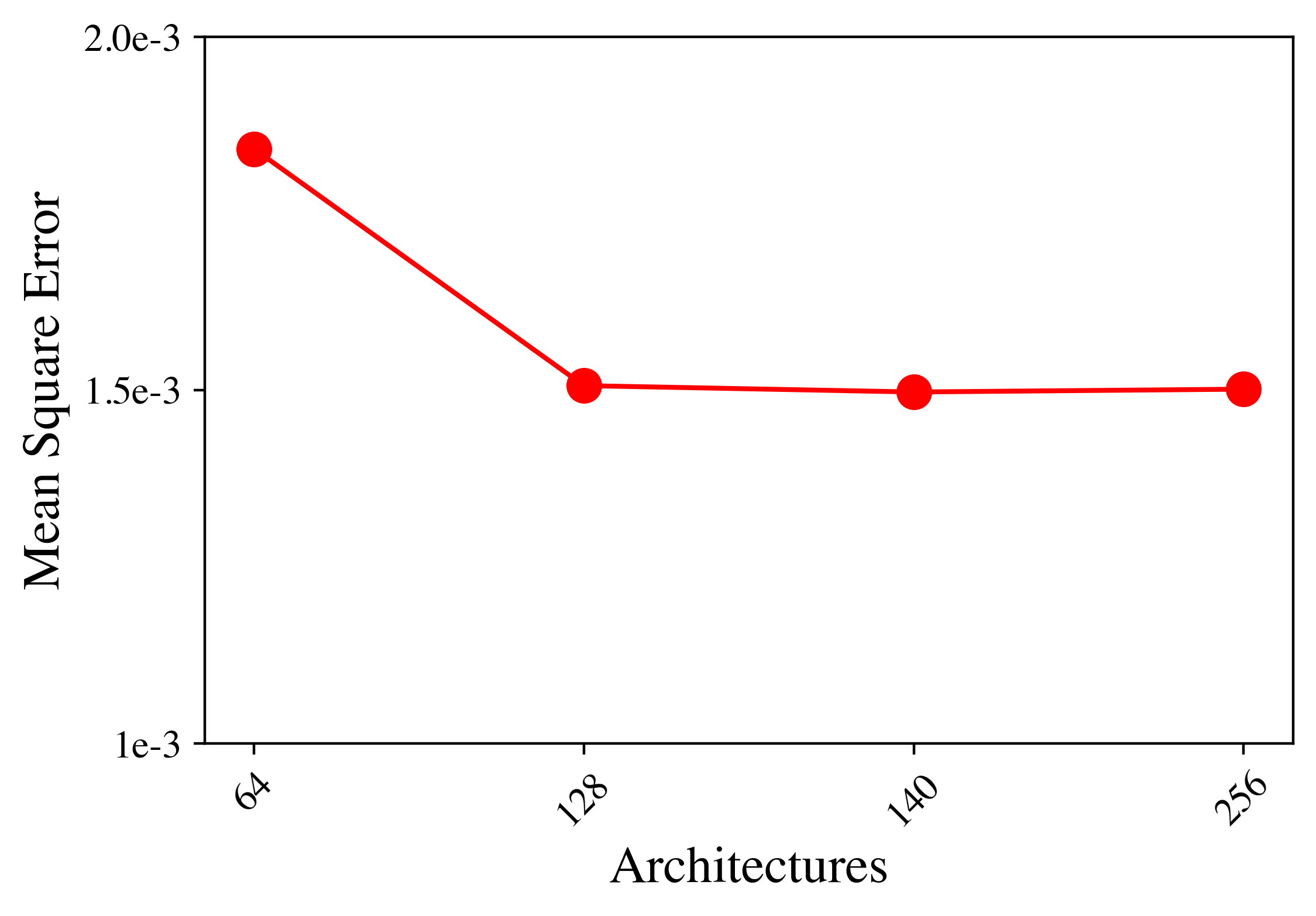}
\caption{Average Mean Square Error of MLPs with different architectures on test data.\\}
\end{subfigure}
\caption{The sensitivity analysis of interaction detection algorithms to the regularization strength~(L1 is set to $5e-6$).}
\label{fig:different architecture l1 5e-6}
\end{figure}

\begin{figure}[hbt!]
\centering
\begin{subfigure}[h]{0.4\linewidth}
\centering
\includegraphics[width=\linewidth]{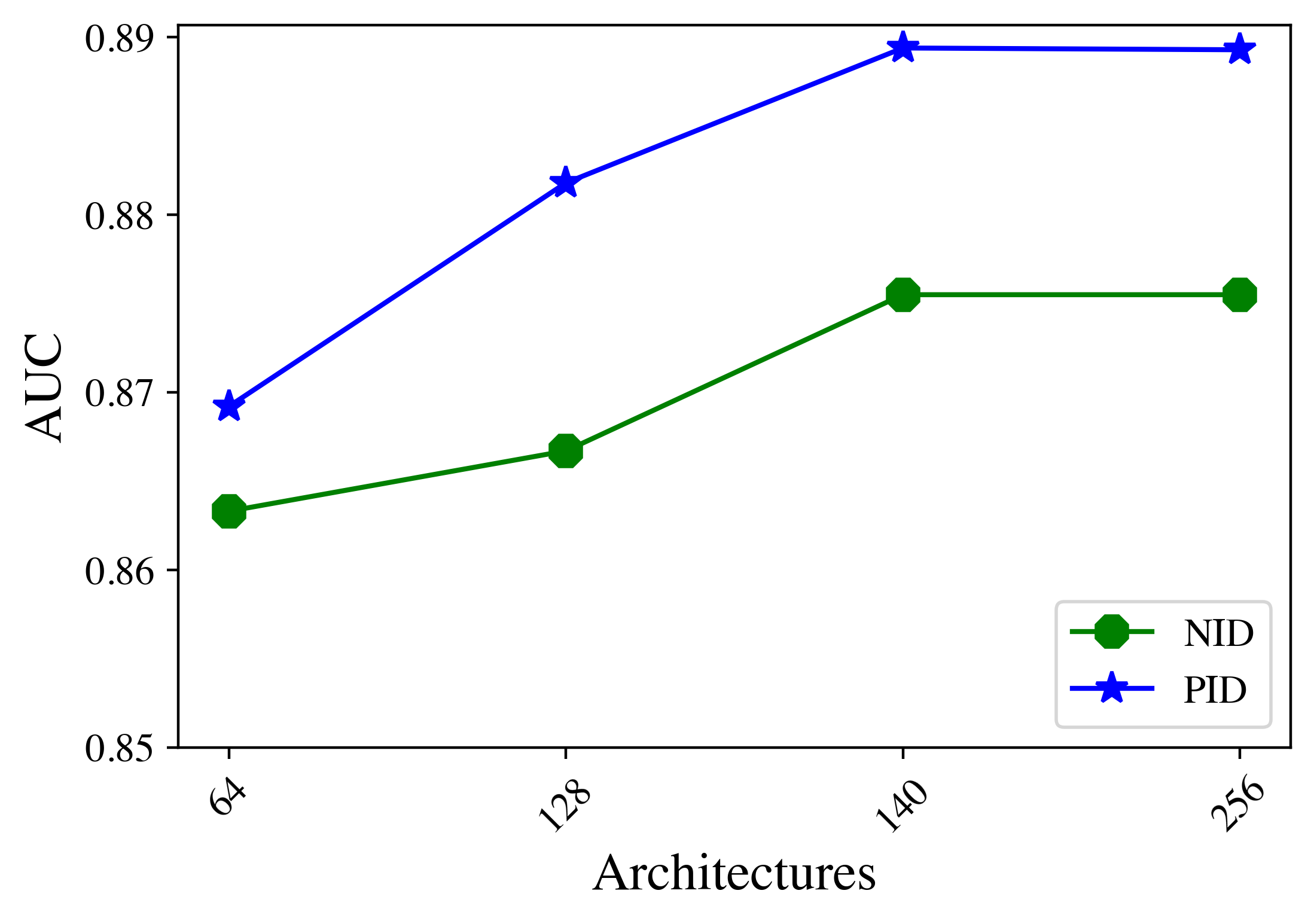}
\caption{Average AUCs of pairwise interaction detected by NID and PID using MLPs with different architectures.}
\end{subfigure}\hspace{0.05\textwidth}
\begin{subfigure}[h]{0.4\linewidth}
\centering
\includegraphics[width=\linewidth]{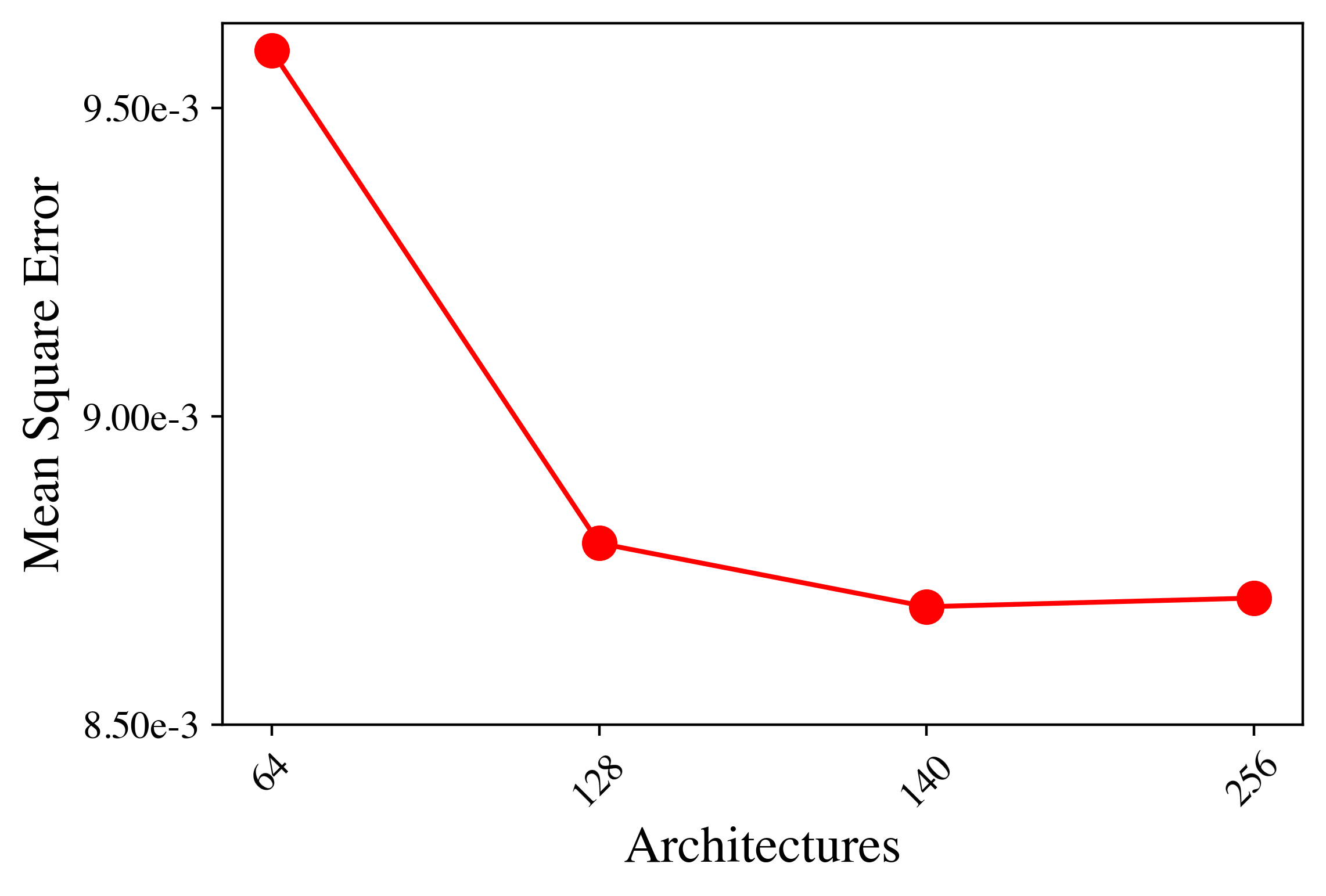}
\caption{Average Mean Square Error of MLPs with different architectures on test data.\\}
\end{subfigure}
\caption{The sensitivity analysis of interaction detection algorithms to the regularization strength~(L1 is set to $5e-4$).}
\label{fig:different architecture l1 5e-4}
\end{figure}
\begin{table}[hbt!]
  \caption{AUC of pairwise interaction strengths proposed by PID and NID on the synthetic functions. The L1 regularization strength is set to $5e-4$ here.}
  \label{tab:result for syn-f l1 5e-4}
  \centering
\begin{tabular}{c|cc}
\toprule
  &NID&PID \\
\hline
$F_{1}(x)$ &$0.898\pm0.0145$ &   $\mathbf{0.915\pm0.0144}$ \\
$F_{2}(x)$ &$0.700\pm0.0419$ &   $\mathbf{0.717\pm0.0349}$ \\
$F_{3}(x)$ &$0.964\pm0.0318$ &   $\mathbf{0.966\pm0.0342}$ \\
$F_{4}(x)$ &$0.928\pm0.0649$ &   $\mathbf{0.938 \pm0.0585}$ \\
$F_{5}(x)$ &$1.000\pm0.0000$ &   $1.000\pm0.0000$ \\
$F_{6}(x)$ &$0.740\pm0.0531$ &   $\mathbf{0.769\pm0.0669}$ \\
$F_{7}(x)$ &$\mathbf{0.807\pm0.0318}$ &   $0.806\pm0.0385$ \\
$F_{8}(x)$ &$0.996\pm0.0085$ &   $\mathbf{0.997\pm0.0084}$ \\
$F_{9}(x)$ &$0.785\pm0.0778$ &   $\mathbf{0.811\pm0.0475}$ \\
$F_{10}(x)$ &$\mathbf{0.937\pm0.0285}$ &  $0.927\pm0.0383$ \\
\hline
average &$0.876\pm0.1033$       &  $\mathbf{0.885\pm0.0954}$  \\
\hline
\end{tabular}
\end{table}

\begin{figure}[hbt!]
    \centering
    \includegraphics[scale=0.25]{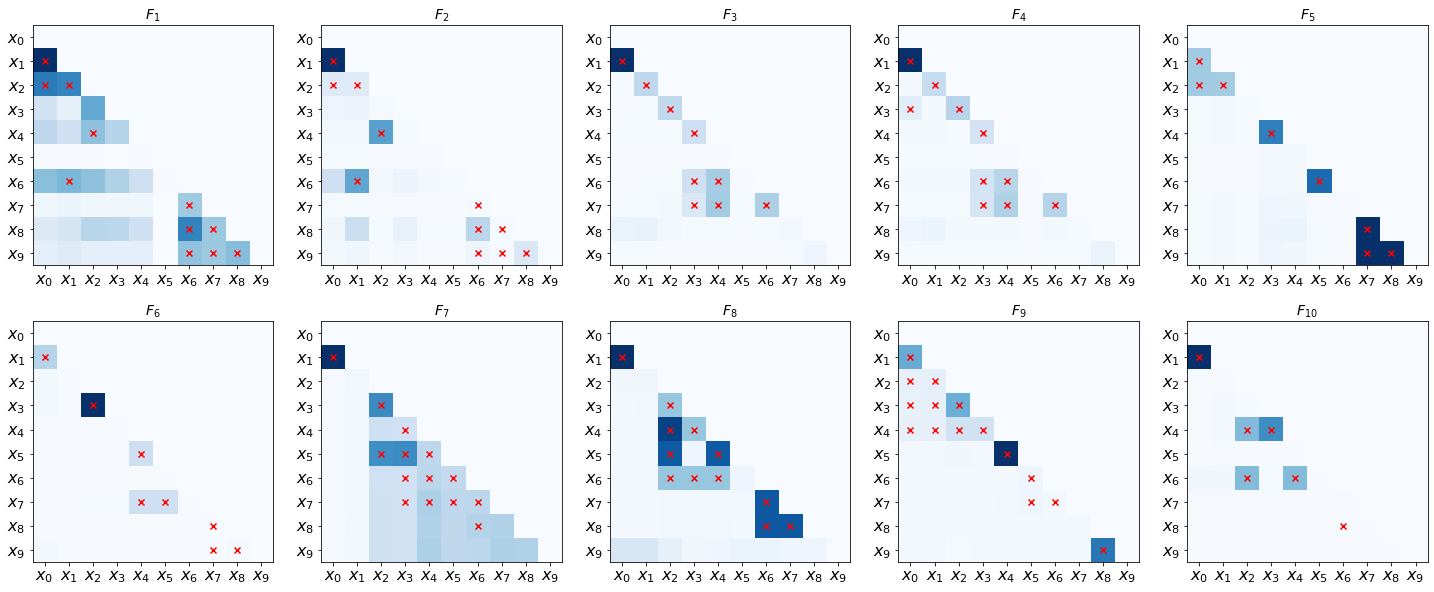}
    \caption{Heat maps of pairwise interaction strengths proposed by our PID corresponding to Table~\ref{table:result for syn-f}. Cross-marks indicate ground truth interactions.~(L1 is set to $5e-4$).}
    \label{fig:heat map l1 5e-4}
\end{figure}

The regularization strength controls the weight sparsity in neural networks. Intuitively, it significantly influences the interaction detection results because it will change the connectivity in networks. Here we change the L1 strength to 5e$-$4 and 5e$-$6, and all other experiment settings are identical. 

Figure~\ref{fig:different architecture l1 5e-6} shows the results using MLP with L1 set to $5e-6$. The average MSE of all MLP models used here is less than $2e-3$ on test data. Similar to Figure~\ref{fig:different architecture default setting}, PID can achieve better performance than NID but the gap is small. Figure~\ref{fig:different architecture l1 5e-4} shows the results using MLP with L1 set to $5e-4$. The average MSE of all MLP models used here is less than $1e-2$ on test data. Comparing Figure~\ref{fig:different architecture l1 5e-4} with Figure~\ref{fig:different architecture default setting}, the mean square error is worse but is acceptable. However, the false discovery rate increases dramatically. To 
better understand the impact of regularization strength, we further analyze the MLP of 140-100-60-20 architecture. Similar to Figure~\ref{fig:heat map of syn}, we plot the heat map in Figure~\ref{fig:heat map l1 5e-4}, and the detailed results are shown in Table~\ref{tab:result for syn-f l1 5e-4}. Comparing Table~\ref{tab:result for syn-f l1 5e-4} with Table~\ref{table:result for syn-f}, both the performances of PID and NID dropped. Moreover, the AUCs of $F_6$ and $F_9$ dropped more than 0.1. Here we provide a detailed case study for MLPs trained on synthetic dataset $F_6$. Comparing Figure~\ref{fig:heat map l1 5e-4} with Figure~\ref{fig:heat map of syn}, it should be noted that the interaction strength between $\{x_7,x_8,x_9\}$ is very small~(near 0 actually). As ~\cite{NID} Appendix I points out, in synthetic dataset $F_6$, $\{x_7,x_8,x_9\}$ can be approximated as
\be
\sqrt{x^2_7+x^2_8+x^2_9}\approx c + x^2_7+x^2_8+x^2_9. \nonumber
\ee

In \cite{NID}, the authors show that $\{x_7,x_8,x_9\}$ are modeled as spurious
main effects in the MLP-M~(the MLP-M is an MLP with optional univariate networks, which details can be found in \cite{NID} Figure 2). Here we hypothesize that, under strong regularization strength, they are also modeled as spurious main effects in MLPs. Figure~\ref{fig:weights dist of diff neurons l1 5e-4} shows the weight statistics of the magnitudes of $\mathbf{W}^{(1)}_{7,:}$, $\mathbf{W}^{(1)}_{8,:}$, and $\mathbf{W}^{(1)}_{9,:}$ of an MLP trained on $F_6$. There is a similar pattern between Figure~\ref{fig:weights dist of diff neurons l1 5e-4} and Figure~\ref{fig:stat of x8andx9 in f3}. Similar to Figure~\ref{fig:weights dist of diff neurons}, we further plot the weights of input features to the representative neurons corresponding to the peaks in Figure~\ref{fig:representative neurons l1 5e-4}. We remark that, for all these neurons corresponding to peaks in Figure \ref{fig:weights dist of diff neurons l1 5e-4}, they share a similar pattern. Therefore, we show only one of their statistics for illustrative purposes. In Figure~\ref{fig:representative neurons l1 5e-4}, we select neuron 5, 131, and 19 for $\mathbf{W}^{(1)}_{7,:}$, $\mathbf{W}^{(1)}_{8,:}$, and $\mathbf{W}^{(1)}_{9,:}$, respectively. From Figure~\ref{fig:representative neurons l1 5e-4}, it can be seen that MLP do not model the interaction $\{x_7,x_8,x_9\}$. Instead, they are modeled as spurious main effects.


\begin{figure}[hbt!]
\centering
\begin{subfigure}[h]{0.32\linewidth}
\centering
\includegraphics[width=\linewidth]{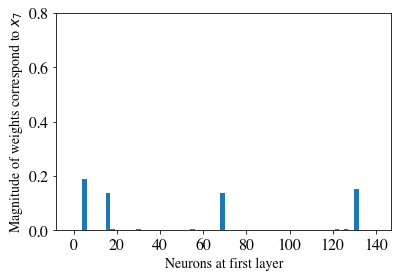}
\end{subfigure}
\begin{subfigure}[h]{0.32\linewidth}
\centering
\includegraphics[width=\linewidth]{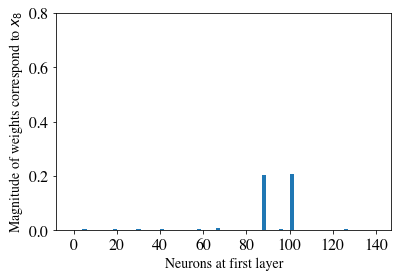}
\end{subfigure}
\begin{subfigure}[h]{0.32\linewidth}
\centering
\includegraphics[width=\linewidth]{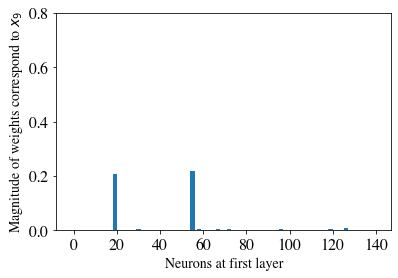}
\end{subfigure}
\caption{Statistics of the magnitudes of weights corresponding to $x_7$, $x_8$ and $x_9$ at different neurons of the first layer~(the MLP is trained on $F_6$ with L1 regularization strength set to $5e-4$).}
\label{fig:weights dist of diff neurons l1 5e-4}
\end{figure}

\begin{figure}[hbt!]
\centering
\begin{subfigure}[h]{0.32\linewidth}
\centering
\includegraphics[width=\linewidth]{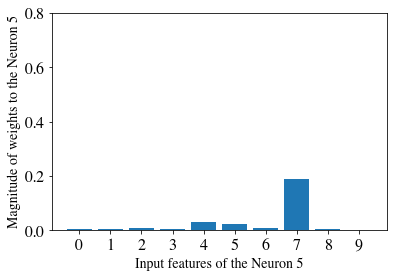}
\end{subfigure}
\begin{subfigure}[h]{0.32\linewidth}
\centering
\includegraphics[width=\linewidth]{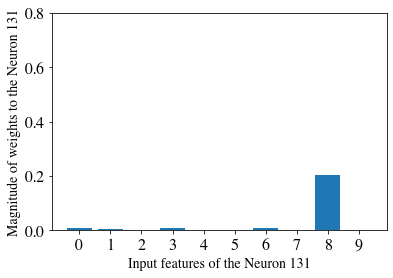}
\end{subfigure}
\begin{subfigure}[h]{0.32\linewidth}
\centering
\includegraphics[width=\linewidth]{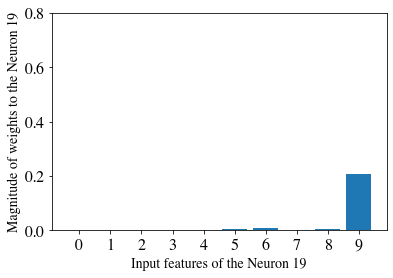}
\end{subfigure}
\caption{The magnitude of weights corresponding to different input features at the selected representative neurons of $x_7,x_8$ and $x_9$ ~(L1 is set to $5e-4$).}
\label{fig:representative neurons l1 5e-4}
\end{figure}

In conclusion, both NID and PID are insensitive to the architecture of MLPs and both of them are sensitive to the regularization strength. A detailed case study for the impact of regularization strength is shown in Figure~\ref{fig:heat map l1 5e-4}, Figure~\ref{fig:weights dist of diff neurons l1 5e-4}, and Figure \ref{fig:representative neurons l1 5e-4}. This suggests that we should carefully choose the regularization strength. From Figure~\ref{fig:different architecture default setting}, Figure~\ref{fig:different architecture l1 5e-4}, and Figure~\ref{fig:different architecture l1 5e-6},
PID always achieves better performance. Also, we observe PID is more resilient to changes in regularization strength. Generally speaking, interaction detection algorithms have better AUC when the MLP has better performance. It makes sense that, when the MLP fits the true distribution, the interactions encoded in the networks are more accurate.

\section{Details for the Automatic Feature Engineering Experiments}
\subsection{Experiment Setting}
\label{appendix: Details for automatic feature engineering experiments}

\begin{table}[hbt!]
\centering
\caption{Statistics of datasets. ``\# Dense''and ``\# Sparse'' are the number of numerical features and the number of categorical features, respectively. ``\# Samples'' is total available samples in each dataset.}
\label{tab:fe_dataset}
\begin{tabular}{|c|c|c|c|}
\hline
\multicolumn{1}{|c|}{\multirow{2}{*}{Dataset}} & \multicolumn{1}{c|}{\multirow{2}{*}{\#Samples}} & \multicolumn{2}{c|}{\# Features}                               \\ \cline{3-4} 
\multicolumn{1}{|c|}{}                      & \multicolumn{1}{c|}{}                           & \multicolumn{1}{l|}{\# Dense} & \multicolumn{1}{l|}{\# Sparse} \\ \hline
Amazon Employee    &   32769          &        0        &              9        \\
Higgs Boson      &  98050          &    28         &        0                        \\
Creditcard       & 284807        &       30   &        0                        \\
Spambase        &   4601      &          57          &      0         \\
Diabetes         &     768   &        8             &        0         \\
\hline
\end{tabular}
\end{table}

We perform most of the experiments on five open-source tabular datasets from different
domains: \textbf{Amazon Employee}\footnote[2]{https://www.kaggle.com/c/amazon-employee-access-challenge}, \textbf{Higgs Boson}\footnote[3]{https://archive.ics.uci.edu/ml/datasets/HIGGS}, \textbf{Creditcard} \footnote[4]{https://www.openml.org/d/1597}, \textbf{Spambase}\footnote[5]{https://archive.ics.uci.edu/ml/datasets/spambase} and \textbf{Diabetes}\footnote[6]{https://www.openml.org/d/37}. For the ease to reproduce our results, we use OpenML \cite{feurer2019openml} to obtain all these datasets and adopt standard cross validation provided by OpenML. The statistics of datasets we used in Section~\ref{exp: afe} is described in Table~\ref{tab:fe_dataset}.

The MLPs for NID and
PID have architectures of 256-128-64 first-to-last hidden layer sizes, and they are trained
with learning rate of $5e-3$, batchsize of 100, and the Adam optimizer. As pointed out in Appendix~\ref{appendix:sentivity of syn exp}, the regularization strength significantly influences the results of NID and PID. We tune the L1 regularization strength with a search space $[1e-6, 1e-1]$ for each dataset. The early stopping round is set to 20 to prevent overfitting.

The synthetic feature $\mathbf{x}_{\mathcal{L}_i}$ is created by explicitly crossing sparse features
indexed in $\mathcal{L}_i$. If interaction $\mathcal{L}_i$ involves dense features, we bucketize the dense features before crossing them. The bucket size is set to 100 across all experiments. Let $|\mathcal{L}_i|=t$ and $\{0,...,t-1\}$ is the interaction candidate specified by $\mathcal{L}_i$. A synthetic feature $\mathbf{x}_{\mathcal{L}_i}$ is an $t$-ary Cartesian product among $t$ features, which means $\mathbf{x}_{\mathcal{L}_i}$ takes on all possible values in $\{(x_1,...,x_t)|\forall x_i \in \mathbf{x}_i, i=0,...,t-1\}$.

Concerning the cardinality of synthetic features can be extremely large, yet
many combinations do not exist in the training data, we limit the order of crossing features up to 4 over all five datasets. For sparse categorical features, like CatBoost~\cite{dorogush2018catboost}, we apply target encoding to make them applicable to the random forest.

 We run five trials of PID and NID on each dataset to obtain five different sets of top ten interactions. For each set of top ten interactions, we construct synthetic features and integrate them with original input features, and then we split the concatenated data into five folds. Subsequently, five random forest models are trained and evaluated with each fold given a chance to be the test set. Totally, we trained 25 random forest models on each dataset and removed two models with the highest and the lowest performance. We implement the random forest via LightGBM~\cite{ke2017lightgbm}. 
The hyperparameters of random forest is summerized in Table \ref{tab:hyperparameters for random forest}.

\begin{table}[hbt!]
\centering
\caption{Hyperparameters of the random forest.}
\label{tab:hyperparameters for random forest}
\begin{tabular}{|c|c|} 
\hline
Name                                          & Value                    \\
\hline
early\_stopping\_rounds & 50 \\
num\_boost\_round                             & 5000                     \\
learning\_rate                                & 0.05                     \\
lambda\_1                                     & 0.2                      \\
lambda\_2                                     & 0.2                      \\
bagging\_raction                              & 0.85                     \\
bagging\_req                                  & 3                        \\
\hline
\end{tabular}
\end{table}

\subsection{Additional Experiment Results}
\label{appendix: additional experiment results for autofe exp}

\begin{table}[hbt!]
\caption{Interaction order statistics.}
\label{tab: interaction order statistics}
\begin{tabular}{ccccccc}
\toprule
Method                                   &      & Amazon Employee & Higgs Boson & Creditcard & Spambase & Diabetes \\
\hline
\multicolumn{1}{c}{\multirow{3}{*}{NID}} & max  & $4.00\pm0.00$ & $4.00\pm0.00$ & $4.00\pm0.00$ &   $4.00\pm0.00$        &     $3.60\pm0.80$       \\
\multicolumn{1}{c}{}                     & mean & $3.30\pm0.06$ & $2.50\pm0.11$ & $2.70\pm0.17$ &   $2.62\pm0.12$       &   $2.30\pm0.17$  \\
\multicolumn{1}{c}{}                     & min  & $2.00\pm0.00$ & $2.00\pm0.00$ & $2.00\pm0.00$ &   $2.00\pm0.00$         & $2.00\pm0.00$         \\
\hline
\multirow{3}{*}{PID}                     & max  & $4.00\pm0.00$ & $3.80\pm0.40$ & $4.00\pm0.00$ &   $4.00\pm0.00$       &        $4.00\pm0.00$  \\
                                         & mean & $3.30\pm0.14$ & $2.64\pm0.20$ & $2.84\pm0.31$ & $2.94\pm0.24$         &  $3.48\pm0.35$        \\
                                         & min  & $2.00\pm0.00$ & $2.00\pm0.00$ & $2.00\pm0.00$ & $2.00\pm0.00$        & $2.40\pm0.49$\\         
\hline
\end{tabular}
\end{table}

The statistics of detected interaction orders by PID and NID are shown in Table \ref{tab: interaction order statistics}. Interaction orders are averaged over 5 folds of cross-validation.

Here we present the case study for the ``Amazon Employee'' dataset in Table \ref{tab: case study of amazon employee} and Table \ref{tab: case study of amazon employee,NID}. The main reasons for choosing ``Amazon Employee'' are as follows: first, it is a dataset used for Kaggle challenges and, thus, the top solution is available. Second, the key technique in the top solution is to construct synthetic features for 2-order and 3-order interactions, so we can compare our detected interactions against the best hand-crafted interactions.

\begin{table}[hbt!]
\centering
\caption{Top ten interaction candidates proposed by PID for Amazom Employee dataset.}
\label{tab: case study of amazon employee}
\begin{tabular}{cc}
\toprule
Interaction Candidates & Interaction Strength\\
\hline
\{\texttt{RESOURCE, MGR\_ID, ROLE\_FAMILY\_DESC}\} & 2.206 \\
\{\texttt{RESOURCE, MGR\_ID}\} & 1.456 \\
\{\texttt{RESOURCE, MGR\_ID, ROLE\_DEPTNAME, ROLE\_FAMILY\_DESC}\} & 1.333 \\
\{\texttt{MGR\_ID, ROLE\_FAMILY\_DESC}\} & 0.418 \\
\{\texttt{RESOURCE, MGR\_ID, ROLE\_DEPTNAME}\} & 0.393 \\
\{\texttt{RESOURCE, MGR\_ID, ROLE\_TITLE, ROLE\_FAMILY}\} & 0.385 \\
\{\texttt{RESOURCE, MGR\_ID, ROLE\_ROLLUP\_2, ROLE\_FAMILY\_DESC}\} & 0.315 \\
\{\texttt{RESOURCE, MGR\_ID, ROLE\_TITLE, ROLE\_FAMILY\_DESC}\} & 0.270 \\
\{\texttt{RESOURCE, MGR\_ID, ROLE\_FAMILY}\} & 0.220 \\
\{\texttt{MGR\_ID, ROLE\_DEPTNAME}\} & 0.190 \\
\hline
\end{tabular}
\end{table}

\begin{table}[hbt!]
\centering
\caption{Top ten interaction candidates proposed by NID for Amazom Employee dataset.}
\label{tab: case study of amazon employee,NID}
\begin{tabular}{cc}
\toprule
Interaction Candidates & Interaction Strength\\
\hline
\{\texttt{RESOURCE, MGR\_ID, ROLE\_FAMILY\_DESC}\} & 26.757 \\
\{\texttt{RESOURCE, MGR\_ID}\} & 22.060 \\
\{\texttt{RESOURCE, MGR\_ID, ROLE\_DEPTNAME, ROLE\_FAMILY\_DESC}\} & 10.423 \\
\{\texttt{MGR\_ID, ROLE\_FAMILY\_DESC}\} & 7.713 \\
\{\texttt{RESOURCE, MGR\_ID, ROLE\_TITLE, ROLE\_FAMILY\_DESC}\} & 2.697 \\
\{\texttt{RESOURCE, MGR\_ID, ROLE\_FAMILY\_DESC, ROLE\_CODE}\} & 2.448 \\
\{\texttt{RESOURCE, ROLE\_FAMILY\_DESC}\} & 2.436 \\
\{\texttt{RESOURCE, MGR\_ID, ROLE\_ROLLUP\_2, ROLE\_FAMILY\_DESC}\} & 2.316 \\
\{\texttt{RESOURCE, MGR\_ID, ROLE\_FAMILY\_DESC, ROLE\_FAMILY}\} & 1.187 \\
\{\texttt{ROLE\_CODE, MGR\_ID, ROLE\_TITLE, ROLE\_FAMILY\_DESC}\} & 1.070 \\

\hline
\end{tabular}
\end{table}

In general, the interaction candidates detected by NID and PID are similar. However, there exists some interaction candidates only detected by PID or NID, respectively. For example, ``\{\texttt{MGR\_ID, ROLE\_FAMILY\_DESC}\}'' are only detected by NID. We note that the scale of the interaction strength proposed by PID and NID are different and only the rankings of interaction candidates are comparable.
From Table \ref{tab: case study of amazon employee}, most of the interaction candidates proposed by PID for Amazon Employee are 3-order interactions. None of the top ranked interactions contain the input feature \texttt{ROLE\_CODE}. This result is consistent with the top solution: ``Transform the data to higher degree features by considering all pairs and triples of the original data ignoring~\texttt{ROLE\_CODE}''\footnote[7]{https://www.kaggle.com/c/amazon-employee-access-challenge/discussion/4838}. In contrast, ``\texttt{ROLE\_CODE}'' are contained in the interaction candidates proposed by NID. And our top ranked interactions are also consistent with the hand-designed synthetic features built from interactions,\footnote[8]{https://www.kaggle.com/c/amazon-employee-access-challenge/discussion/5283} such as \{\texttt{RESOURCE, MGR\_ID}\} corresponding to ``The number of unique resources that a MGR\_ID received requests for''.

\section{Details for High-order Interaction Detection on Image Datasets}
\begin{figure}[hbt!]
  \centering
  \includegraphics[scale=0.4]{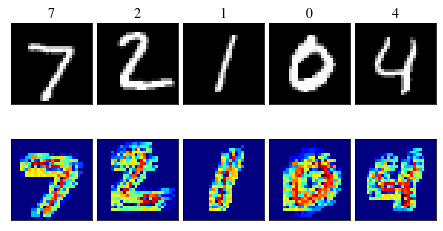}
  \caption{Saliency maps of interaction strength found from applying PID on the CNN trained on MNIST dataset.}
  \label{fig:mnist_pid_res}
\end{figure}
\label{appendix:Details for local interacion detection on image datasets}
The neural network is composed of two convolutional layers of kernel size 5 and stride 1, followed by a max pooling layer and ReLU activation, and ended with a dense layer. The two convolutional layers contain 8 and 16 filters, respectively. It is trained with learning rate of $5e-3$, batchsize of 100, the Adam optimizer, L1 regularization of $5e-4$, and train epochs of 5.

Similar to Figure \ref{fig:fashionmnist_pid_res}, Figure \ref{fig:mnist_pid_res} also shows that PID are capable of detecting high-order interactions that represent object shapes.

\end{document}